\documentclass{article}
\usepackage{textcomp}
\usepackage{amssymb}
\usepackage{algorithm}

\usepackage{algpseudocode}
\usepackage{amsmath}
\usepackage{pifont}
\usepackage{multirow}
\usepackage{xcolor}
\usepackage{graphicx}
\usepackage{adjustbox}
\usepackage{subcaption}
\usepackage{float}  
\usepackage{booktabs}
\UseRawInputEncoding

\newcommand{\cmark}{\textcolor{green!60!black}{\ding{51}}}
\newcommand{\xmark}{\textcolor{red}{\ding{55}}}

\usepackage[preprint]{corl_2025} 

\title{GRoQ-LoCO: Generalist and Robot-agnostic Quadruped Locomotion Control using \\ Offline Datasets}

\vspace{1em}

\author{
  \textbf{Narayanan PP} \quad 
  \textbf{Sarvesh Prasanth Venkatesan} \quad 
  \textbf{Srinivas Kantha Reddy} \quad 
  \textbf{Shishir Kolathaya} \\
  \textnormal{Indian Institute of Science, Bangalore, India} \\
  \texttt{pnkrishnan27@gmail.com, sarveshv219@gmail.com,} \\
  \texttt{srinivas299792458@gmail.com, shishirk@iisc.ac.in}
}

\begin{document}
\maketitle


\begin{abstract}
Recent advancements in large-scale offline training have demonstrated the potential of generalist policy learning for complex robotic tasks. However, applying these principles to legged locomotion remains a challenge due to continuous dynamics and the need for real-time adaptation across diverse terrains and robot morphologies. In this work, we propose GRoQ-LoCO, a scalable, attention-based framework that learns a single generalist locomotion policy across multiple quadruped robots and terrains, relying solely on offline datasets. Our approach leverages expert demonstrations from two distinct locomotion behaviors - stair traversal (non-periodic gaits) and flat terrain traversal (periodic gaits) - collected across multiple quadruped robots, to train a generalist model that enables behavior fusion. Crucially, our framework operates solely on proprioceptive data from all robots without incorporating any robot-specific encodings. The policy is directly deployable on an Intel i7 nuc, producing low-latency control outputs without any test-time optimization. Our extensive experiments demonstrate zero-shot transfer across highly diverse quadruped robots and terrains, including hardware deployment on the Unitree Go1, a commercially available 12kg robot. Notably, we evaluate challenging cross-robot training setups where different locomotion skills are unevenly distributed across robots, yet observe successful transfer of both flat walking and stair traversal behaviors to all robots at test time. We also show preliminary walking on Stoch 5, a 70kg quadruped, on flat and outdoor terrains without requiring any fine tuning. These results demonstrate the potential of offline, data-driven learning to generalize locomotion across diverse quadruped morphologies and behaviors.


\end{abstract}


\section{Introduction}\label{sec:intro}

Generalization is a central challenge in legged locomotion control. Robust controllers must not only produce stable and efficient motions but also adapt to new terrains, disturbances, and robot designs without the need for retraining. Achieving such generalization would enable legged robots to move out of controlled labs and operate reliably in the real world.

In robotic manipulation, \textit{large-scale offline learning} has enabled this kind of generalization. Models like RT-1 and RT-2 \cite{brohan2022rt1, brohan2023rt2}, and efforts like Open X-Embodiment~\cite{openx2024} have demonstrated how diverse, pre-collected datasets can be used to train generalist policies capable of solving hundreds of tasks across different robotic arms—without online interaction. These methods rely on \textit{multi-task imitation learning}, foundation models, and scalable architectures that unify diverse behaviors across embodiments \cite{openx2024, bousmalis2024robocat}.

Bringing these offline learning principles to \textit{legged locomotion}, however, remains underexplored. Locomotion differs fundamentally from manipulation: it requires \textit{continuous control}, real-time adaptation to dynamic environments, and often lacks clearly segmented tasks or episodic resets. Still, the benefits of offline learning—scalability, safety, and reusability—make it a compelling direction for legged robotics, where online exploration is risky and expensive.

Some early efforts have begun applying offline learning to locomotion \cite{reske2021imitation, pmlr-v270-huang25a,mothish2024birodiff,omahoney2024offline}. \textit{DiffuseLoco} \cite{pmlr-v270-huang25a} showed that diffusion models trained on demonstration data can learn diverse gait patterns and enable zero-shot sim-to-real transfer. However, these were confined to a \textit{single robot morphology}, and the scope of behaviors was limited. Other sequence modeling approaches like Decision Transformers~\cite{bousmalis2024robocat,gajewski2024dtmanip,dong2025rmkv} have shown promise in manipulation but have not been widely adopted for locomotion—especially in multi-embodiment settings.

In contrast, \textit{deep reinforcement learning (RL)} has been the dominant paradigm for learning legged locomotion policies. It has enabled \textit{agile behaviors} such as trotting, jumping, and terrain traversal~\cite{hwangbo2019learning,kumar2021rma,xue2021locotransformer,hoeller2023anymal,wang2023amp,long2024hybrid, kim2024learning}. However, RL methods typically require \textit{large-scale online interaction}, \textit{task-specific reward engineering}, and \textit{carefully tuned simulation environments}. As a result, they often produce \textit{specialized policies} that generalize poorly to new robot morphologies or unseen environments without extensive fine-tuning.

Several RL-based approaches have tried to address this by \textit{explicitly modeling morphology variation}\cite{feng2022genloco,bohlinger2024one}. \textit{GenLoco}~\cite{feng2022genloco} introduced morphology randomization during RL training, enabling a single policy to generalize across different quadrupeds such as A1, Mini Cheetah, and Sirius but only for \textit{velocity tracking on flat terrain}. \textit{MorAL}~\cite{luo2024moral} added an adaptive module to infer robot dynamics implicitly, improving generalization, but still relied on \textit{online RL} and lacked the ability to capture \textit{multiple distinct locomotion behaviors} within a single model.


We introduce \textit{GRoQ-LoCO}, a scalable offline learning framework that unifies locomotion control across different terrains and robot designs. Our core insight is that \textbf{dataset diversity} both in \textit{robot morphologies and locomotion behaviors} is essential for generalization.
GRoQ-LoCO is trained on expert demonstrations of \textit{periodic gaits (flat terrain)} and \textit{non-periodic traversal (stairs)} collected from multiple quadruped robots. It operates directly on \textit{proprioceptive inputs}, without any morphology encoding or post-training optimization. GRoQ-LoCO demonstrates \textit{strong generalization} across both behaviors and embodiments. 

\begin{figure*}[t]
    \centering
    \includegraphics[width=0.99\textwidth]{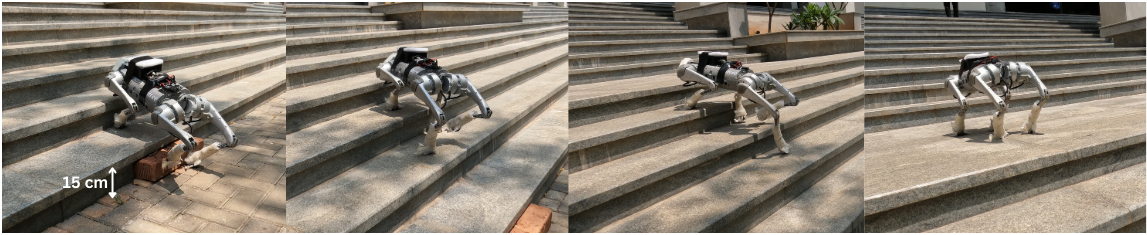}
    \vspace{1mm}
    \caption{Frames of Go1 robot traversing 15cm stairs. The policy showed zero shot transfer to Go1 for flat terrains and staircases.}
    \label{fig:go1robot}
\end{figure*}

The key contributions of our work include:
\begin{itemize}
\item \textbf{A Generalist Locomotion Controller:} We develop a single policy that controls multiple distinct quadrupedal robots without requiring robot-specific information.
\item \textbf{Offline Multi-Behavior Learning:} We demonstrate that purely offline training on diverse motion data produces a policy with periodic gaits and multi-terrain traversibility.
\item \textbf{Zero-Shot Transfer and Robustness:} Our framework achieves strong zero-shot transfer across diverse quadruped robots and terrains, including hardware deployment on commercial platforms like the Unitree Go1 (see Fig. \ref{fig:go1robot}) and the Stoch 5, without requiring fine-tuning.
\end{itemize}

\begin{figure}[t]
    \centering
    \includegraphics[width=0.9\linewidth]{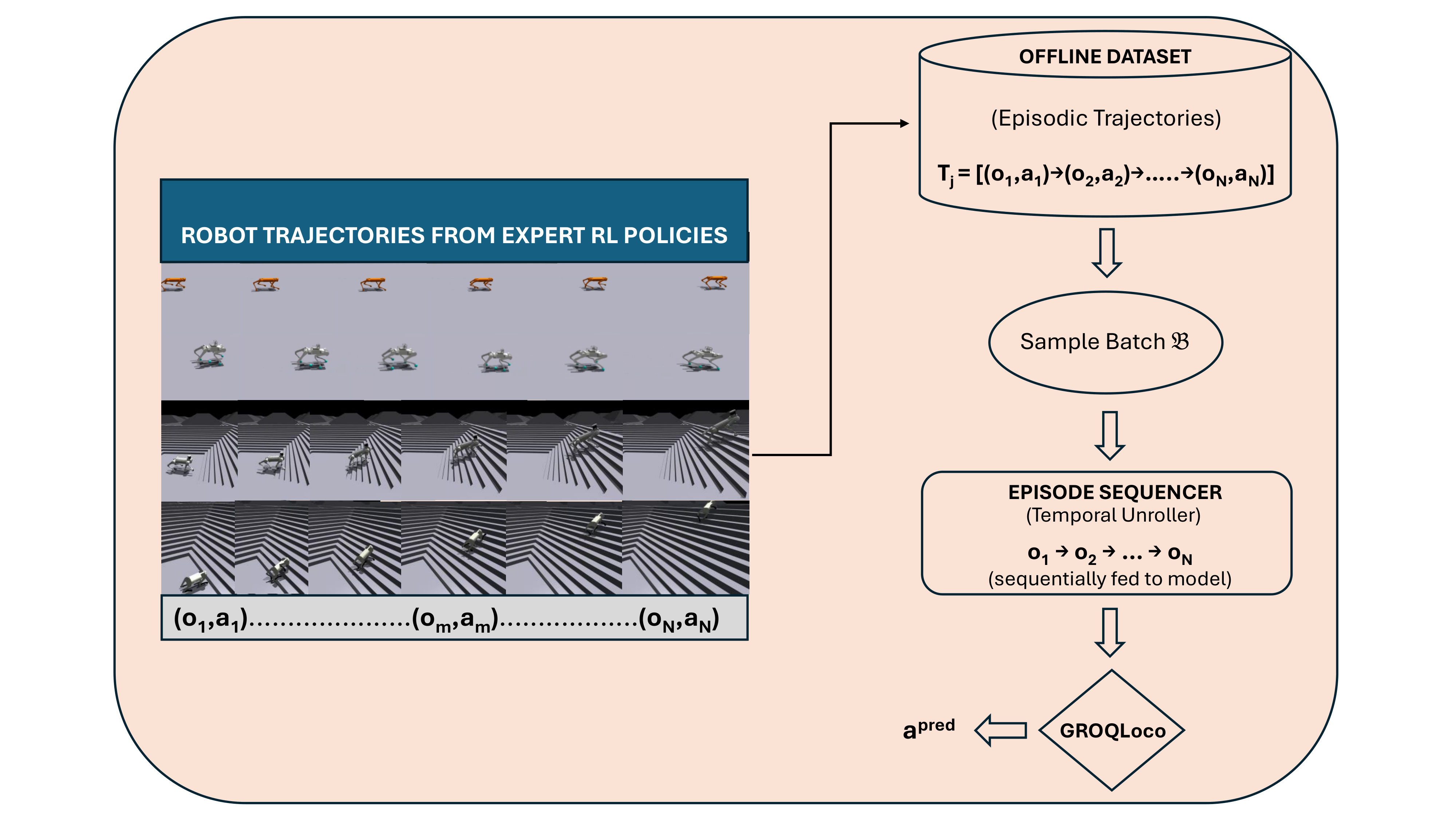}
    \caption{Offline data generation pipeline used in \textbf{GROQLoco}, illustrating trajectory collection from expert RL policies on diverse terrains and robot morphologies.}
    \label{fig:image2_datagen}
\end{figure}

\begin{figure}[t]
    \centering
    \includegraphics[width=0.85\linewidth]{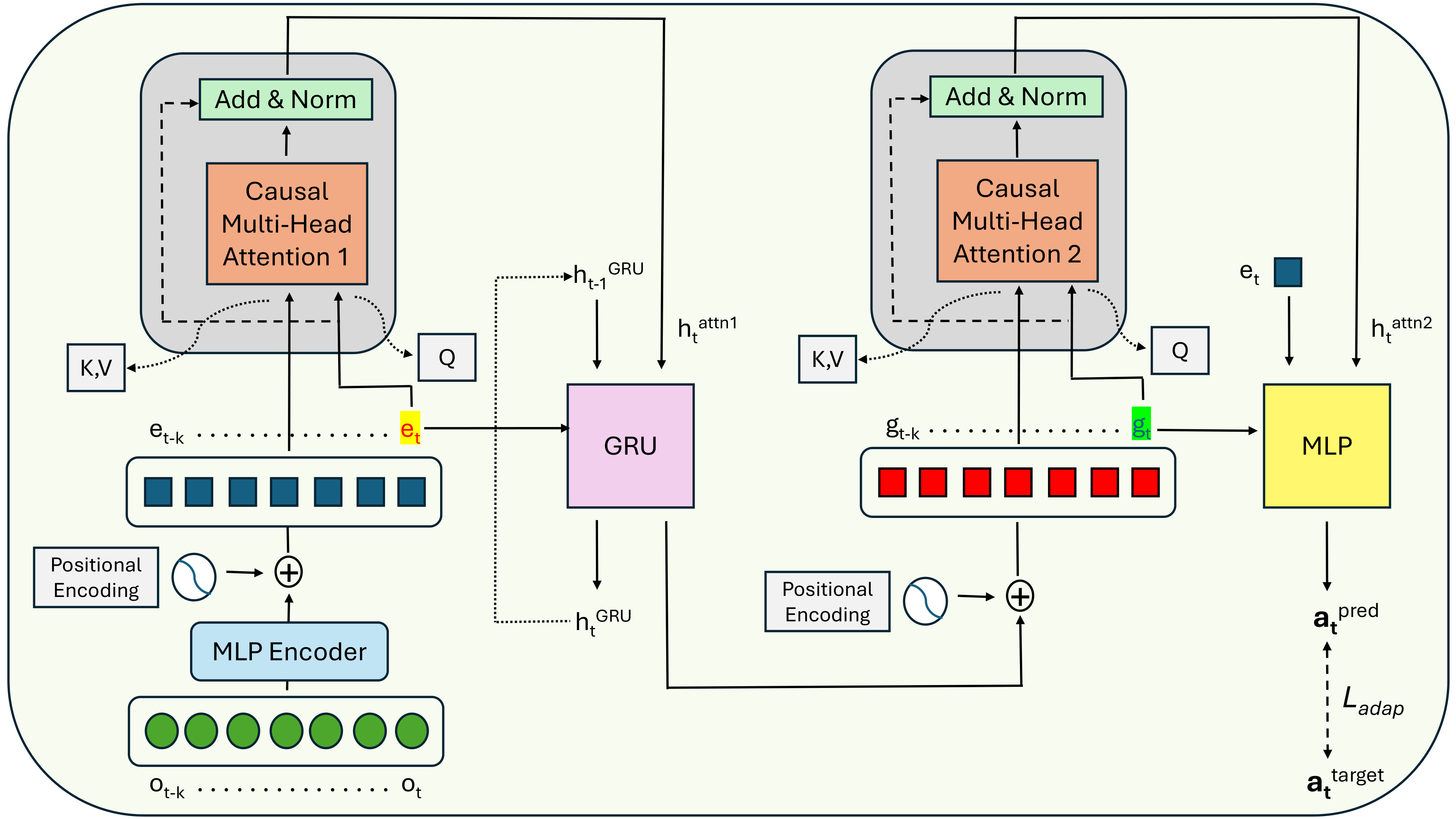}
    \caption{Model architecture of \textbf{GRoQ-LoCO}, showing the sequential processing pipeline with observation encoding, causal attention, GRU-based temporal modeling, and MLP action prediction.}
    \label{fig:image1_model}
\end{figure}

\section{Methodology}

In this section, we present our data collection strategy, the architecture of the proposed behavior cloning policy, the adaptive loss formulation, and the training procedure. Our approach enables a single policy to learn different locomotion behaviors, namely, stable cyclic gaits on flat terrain and stair-climbing on rough terrain, across a diverse set of quadruped robots. We also outline the evaluation of zero-shot generalization to new robots unseen during training.

\subsection{Data Collection and Expert Demonstrations}

We collect expert demonstrations in simulation on select quadruped platforms, including Unitree B2, Go1, Aliengo, and Stoch3 (Refer Fig. \ref{fig:image2_datagen}). Additional platforms such as Unitree B1, X30, Lite3 and Stoch5 are included primarily for evaluating zero-shot generalization and cross-morphology transfer. Two specialized locomotion controllers are used to generate expert trajectories: \begin{itemize} 
\item \textbf{Flat-Ground Controller}: A periodic gait controller designed for stable and efficient locomotion on flat terrain. 

\item \textbf{Rough-Terrain Controller}: A non-periodic controller tailored for stair climbing and traversal of uneven terrain. 

\end{itemize}

Each expert trajectory $\tau$ consists of a sequence of observations $\{o_0, o_1, \dots, o_T\}$ and corresponding expert actions $\{a_0, a_1, \dots, a_T\}$, where each observation $o_t$ includes:
\begin{equation}
o_t = \left[ q_t, \dot{q}_t, a_{t-1}, a_{t-2}, g_t, \omega_t, v^{\text{cmd}}_t \right]^T
\end{equation}
Here, $q_t$ denotes joint positions, $\dot{q}_t$ joint velocities, $a_{t-1}, a_{t-2}$ are previous actions, $g_t$ the gravity-aligned vector, $\omega_t$ the angular velocity, and $v^{\text{cmd}}_t$ the commanded linear and angular velocities $(v_x, v_y, \omega_z)$. No robot-specific identifiers are used; policies are trained purely from proprioceptive and command data to encourage generalization across robots.



\subsection{Policy Architecture}

Our behavior cloning policy is composed of four principal modules: an observation encoder, an attention-enhanced recurrent core, a secondary attention module over the recurrent history, and a final multi-layer perceptron (MLP) for action prediction (Fig. \ref{fig:image1_model}). 

\textbf{Observation Encoder:}
At each timestep $t$, the observation $o_t \in \mathbb{R}^{d_{\text{obs}}}$ is embedded into a latent space via:
\[
\mathbf{e}_t = \mathrm{LayerNorm}(\mathrm{ELU}(\mathbf{W}_e \mathbf{o}_t + \mathbf{b}_e)),
\]
where $\mathbf{W}_e, \mathbf{b}_e$ are learnable parameters.

\paragraph{Positional Embeddings}
To inject temporal ordering information without assuming a fixed history size, we employ fixed sinusoidal positional embeddings:
\[
PE(t, 2i) = \sin\left( \frac{t}{10000^{2i/d_{\text{emb}}}} \right), \quad
PE(t, 2i+1) = \cos\left( \frac{t}{10000^{2i/d_{\text{emb}}}} \right)
\]
These embeddings are added to observation and GRU histories before attention operations.

\textbf{Attention over Observation History:}
The encoded observations $\{e_{t-k}, \dots, e_t\}$ are stacked with positional encoding and processed by a multi-head attention layer:
\[
\mathbf{h_t}^{\text{attn1}} = \text{MHA}_{\text{attn1}}\left( [e_{t-k} + PE, \dots, e_t + PE] \right)
\]

where MHA denotes multi-head self-attention. The most recent $k$ outputs are aggregated.


\textbf{GRU Memory:}
The encoded observation $\mathbf{e}_t$ is concatenated with the attended context $\mathbf{h}_t^{\text{attn1}}$ and passed into a GRU:
\[
\mathbf{g}_t, \mathbf{h}_t^{\text{GRU}} = \mathrm{GRU}([\mathbf{e}_t; \mathbf{h}_t^{\text{attn1}}], \mathbf{h}_{t-1}^{\text{GRU}}).
\]

\textbf{Attention over GRU History:}
The GRU outputs over time $\{\mathbf{g}_{t-k}, \dots, \mathbf{g}_t\}$ are again stacked, positional embeddings added, and processed through a second multi-head attention module:
\[
\mathbf{h}_t^{\text{attn2}} = \text{MHA}\left( [\mathbf{g}_{t-k} + PE, \dots, \mathbf{g}_t + PE] \right)
\]
\textbf{Action Head:}
The final action is computed by feeding the concatenation of $\mathbf{e}_t$, $\mathbf{g}_t$, and $\mathbf{h}_t^{\text{attn2}}$ into an MLP:
\[
\mathbf{a}_t = \mathrm{MLP}([\mathbf{e}_t; \mathbf{g}_t; \mathbf{h}_t^{\text{attn2}}]).
\]



\subsection{Adaptive Loss for Behavior Cloning}

An adaptive loss is employed instead of a standard MSE. Given the predicted action $\hat{\mathbf{a}}_t$ and expert action $\mathbf{a}_t$, the loss is defined as:
\[
\mathcal{L}_{\text{adaptive}} = \frac{1}{T} \sum_{t=1}^T \left( \exp(-\log \sigma) \cdot \delta^2 \log\left(1 + \left(\frac{\hat{\mathbf{a}}_t - \mathbf{a}_t}{\delta}\right)^2\right) + \log \sigma \right),
\]
where $\sigma$ is a learnable parameter per action dimension, and $\delta$ is a fixed scaling hyperparameter (e.g., $\delta=0.5$). $T$ denotes the number of timesteps within each truncated sequence (e.g., $T=20$ during training with truncated BPTT). This loss behaves like a Huber loss with adaptive weighting, allowing important joints to have higher influence during training.

Training is performed using batches of size 400, with truncated BPTT over recurrent states. Hidden states are detached periodically to prevent backpropagation through arbitrarily long sequences. Additional details are present in Appendix C.

\subsection{Training Setup and Details}

Let \(\mathcal{D} = \{\tau_i\}_{i=1}^M\) denote a dataset of \(M\) expert trajectories, where
\[
\tau_i = \{(\mathbf{o}_{i,1}, \mathbf{a}_{i,1}), \dots, (\mathbf{o}_{i,N_i}, \mathbf{a}_{i,N_i})\}
\]
and \(N_i\) is the length of episode \(i\). Define \(N_{\max} = \max_i N_i\). Each \(\tau_i\) is padded to length \(N_{\max}\) with dummy zero vectors and a mask \(\mathbf{m}_{i,t} \in \{0,1\}\) marking valid timesteps.

Training runs for \(E\) epochs. In our setup, one epoch corresponds to sampling a single batch (not a full pass over \(\mathcal{D}\)). At epoch \(e\), we sample a batch \(B\) and extract the padded sequence:
\[
\bigl(\mathbf{o}^{(B)}_t, \mathbf{a}^{(B)}_t, \mathbf{m}^{(B)}_t\bigr), \quad t = 1,\dots,N_{\max}
\]
We process each batch sequentially over time, passing \(\mathbf{o}^{(B)}_t\) and previous hidden state \(\mathbf{h}^{(B)}_{t-1}\) to the model, which outputs predicted action and updated hidden state:
\[
\hat{\mathbf{a}}^{(B)}_t, \mathbf{h}^{(B)}_t = \mathrm{Model}\bigl(\mathbf{o}^{(B)}_t, \mathbf{h}^{(B)}_{t-1}\bigr)
\]
The loss is computed over valid steps:
\[
\mathcal{L}_B = \frac{1}{\sum_t m^{(B)}_t} \sum_{t=1}^{N_{\max}} m^{(B)}_t \, \ell_{\mathrm{adaptive}}\bigl(\hat{\mathbf{a}}^{(B)}_t, \mathbf{a}^{(B)}_t\bigr)
\]
Every \(T_u\) steps (e.g., \(T_u=20\)), we compute gradients:
\[
\nabla_\theta \left( \frac{1}{b} \sum_B \mathcal{L}_B \right)
\]
(where \(b\) is the batch size), update parameters using Adam, and truncate BPTT by detaching hidden states:
\[
\mathbf{h}^{(b)}_t \leftarrow \mathrm{stop\_grad}\bigl(\mathbf{h}^{(b)}_t\bigr) \quad \forall\, t \bmod T_u = 0
\]
To stabilize training, we apply a warmup of \(E_w=50\) epochs where, post-update, all hidden states are reset:
\[
\mathbf{h}^{(b)}_t \leftarrow \mathbf{0} \quad \text{if } e \leq E_w
\]
For \(e > E_w\), hidden states are preserved, enabling continuity across (padded) episodes. This balances TBPTT efficiency with long-horizon memory retention.

\section{Experiments}
\label{sec:result}
We conduct a series of experiments to evaluate how multi-robot and multi-terrain training enables generalist locomotion policies that scale across diverse quadruped embodiments and terrain types. Our setup includes cross-robot training configurations with locomotion skills unevenly distributed across robots. We first study zero-shot transfer and behavior fusion in stair-climbing scenarios, using gait visualizations and step-wise completion tables to capture detailed behavior. We then examine generalization to entirely novel terrains.
Additional analyses, including comparisons with and without explicit robot encodings under identical multi-robot settings, will be provided in the supplementary material.

\begin{table}[t]
\centering

\begin{tabular}{|c|c|c|c|c|c|c|c|}
\hline
\textbf{Setting} & \textbf{Robot} & \textbf{Mode} & \textbf{13 cm} & \textbf{17 cm} & \textbf{21 cm (OOD)} & \textbf{25 cm (OOD)} & \textbf{29 cm (OOD)} \\
\hline
  & Go1     & ZS & \cmark & \cmark & \xmark & \xmark & \xmark \\
  & Stoch5  & ZS & \cmark & \cmark & \cmark & \cmark & \cmark \\
1 & B1      & ZS & \cmark & \cmark & \cmark & \cmark & \cmark \\
  & B2      & FO & \cmark & \cmark & \cmark & \cmark & \cmark \\
  & Aliengo & SO & \cmark & \cmark & \cmark & \xmark & \xmark \\
  & Stoch3  & SO & \cmark & \cmark & \cmark & \cmark & \cmark \\
\hline
  & Go1      & FO & \cmark & \cmark & \xmark & \xmark & \xmark \\
  & Stoch5  & ZS & \cmark & \cmark & \cmark & \xmark & \xmark \\
2 & B1      & ZS & \cmark & \cmark & \cmark & \cmark & \cmark \\
  & B2      & SO & \cmark & \cmark & \cmark & \cmark & \cmark \\
  & Aliengo & ZS & \cmark & \cmark & \xmark & \xmark & \xmark \\
  & Stoch3  & FO & \cmark & \cmark & \cmark & \cmark & \xmark \\
\hline
  & Go1      & FO & \cmark & \cmark & \xmark & \xmark & \xmark \\
  & Stoch5  & ZS & \cmark & \cmark & \cmark & \cmark & \cmark \\
3 & B1      & ZS & \cmark & \cmark & \cmark & \cmark & \cmark \\
  & B2      & ZS & \cmark & \cmark & \cmark & \cmark & \cmark \\
  & Aliengo & SO & \cmark & \cmark & \cmark & \xmark & \xmark \\
  & Stoch3  & SO & \cmark & \cmark & \cmark & \cmark & \cmark \\
\hline
  & Go1      & SO & \cmark & \cmark & \xmark & \xmark & \xmark \\
  & Stoch5  & ZS & \cmark & \cmark & \xmark & \xmark & \xmark \\
4 & B1      & ZS & \cmark & \cmark & \xmark & \xmark & \xmark \\
  & B2      & ZS & \cmark & \xmark & \xmark & \xmark & \xmark \\
  & Aliengo & ZS & \cmark & \cmark & \xmark & \xmark & \xmark \\
  & Stoch3  & FO & \cmark & \cmark & \xmark & \xmark & \xmark \\
\hline
  & Go1     & ZS & \cmark & \cmark & \cmark & \xmark & \xmark \\
  & Stoch5  & ZS & \cmark & \cmark & \cmark & \cmark & \cmark \\
5  & B1      & ZS & \cmark & \cmark & \cmark & \cmark & \cmark \\
  & B2      & SO & \cmark & \cmark & \cmark & \cmark & \cmark \\
  & Aliengo & SO & \cmark & \cmark & \cmark & \cmark & \xmark \\
  & Stoch3  & SO & \cmark & \cmark & \cmark & \cmark & \cmark \\

\hline
\end{tabular}
\vspace{3 mm}  

\caption{Evaluation on stair environments with increasing difficulty (13--29 cm step heights). A checkmark (\cmark) indicates successful climbing of 8 stairs, and a cross (\xmark) indicates failure.}
\label{tab:stair_results}
\end{table}

\begin{figure}[ht]
    \centering
    \begin{subfigure}[b]{0.45\linewidth}
        \centering
        \includegraphics[width=1.1\linewidth]{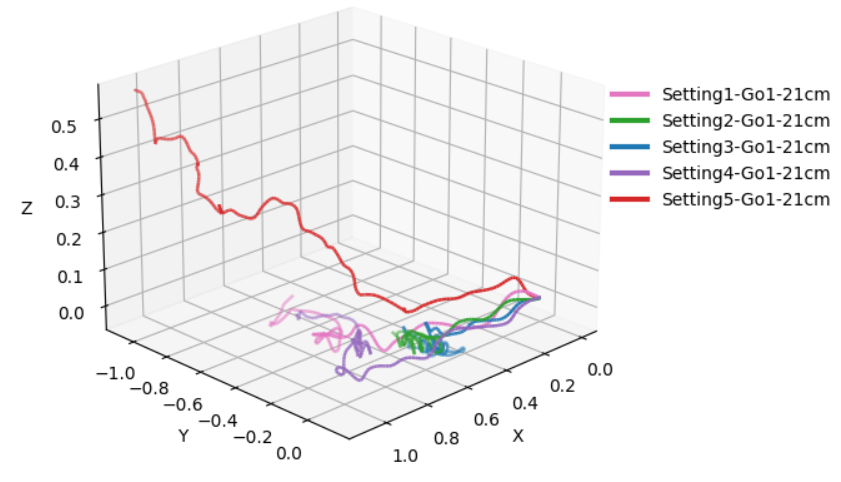}
        \caption{Go1 Body Trajectory Across \textbf{Settings}}
        \label{fig:photo1}
    \end{subfigure}
    \hfill
    \begin{subfigure}[b]{0.45\linewidth}
        \centering
        \includegraphics[width=1.1\linewidth]{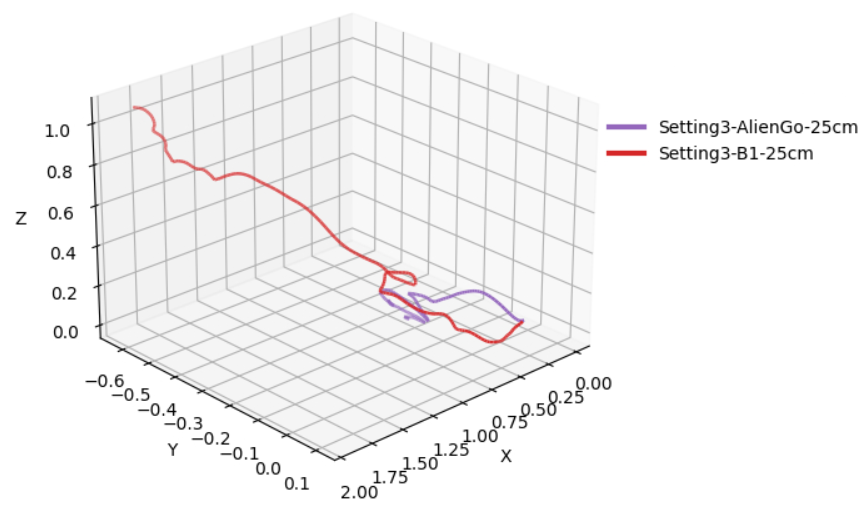}
        \caption{Body Trajectory of \textbf{B1(ZS)} vs \textbf{Aliengo(SO)}}
        \label{fig:photo2}
    \end{subfigure}
    \caption{}
    \label{fig:traj_plot_stairs}
\end{figure}
\vspace{2mm}

\subsection{Cross-Robot and Cross-Terrain Generalization Analysis}
We construct multiple training regimes using combinations of the two locomotion behaviors (periodic gaits and stair traversal) across different robot morphologies. Some configurations include data from both flat and stair policies, while others restrict terrain or robot access to examine generalization. Our goal is to understand how skill and morphology diversity in training data influences a policy's ability to
    \textbf{(a)} Generalize to unseen robots (cross-morphology).
    \textbf{(b)} Transfer learned skills (e.g., from flat to stair).
    \textbf{(c)} Acquire and retain multiple locomotion behaviors simultaneously.
    \textbf{(d)} Handle increasingly difficult Out-of-Distribution (OOD) terrain such as higher stairs.

Our training configurations, each involving a different subset of robots and terrain skills (flat and/or stair). For each configuration, robots are categorized as follows:

\begin{itemize}
    \item \textbf{Zero-Shot (ZS):} The robot is entirely unseen during training (neither flat nor stair).
    
    \item \textbf{Flat Only (FO):} The robot contributed only \textit{flat terrain} data during training.
    
    \item \textbf{Stair Only (SO):} The robot was included in training with \textit{stair terrain} data. 
\end{itemize}

Table~\ref{tab:stair_results} summarizes the results. Flat walking evaluations for the same policies are discussed later in this section to investigate transfer of cyclic motion. All experiments are conducted with a commanded forward velocity of $\mathbf{1\,\mathrm{m/s}}$ along the $x$-axis.

\begin{figure}[t]
    \centering
    \includegraphics[width=0.8\linewidth]{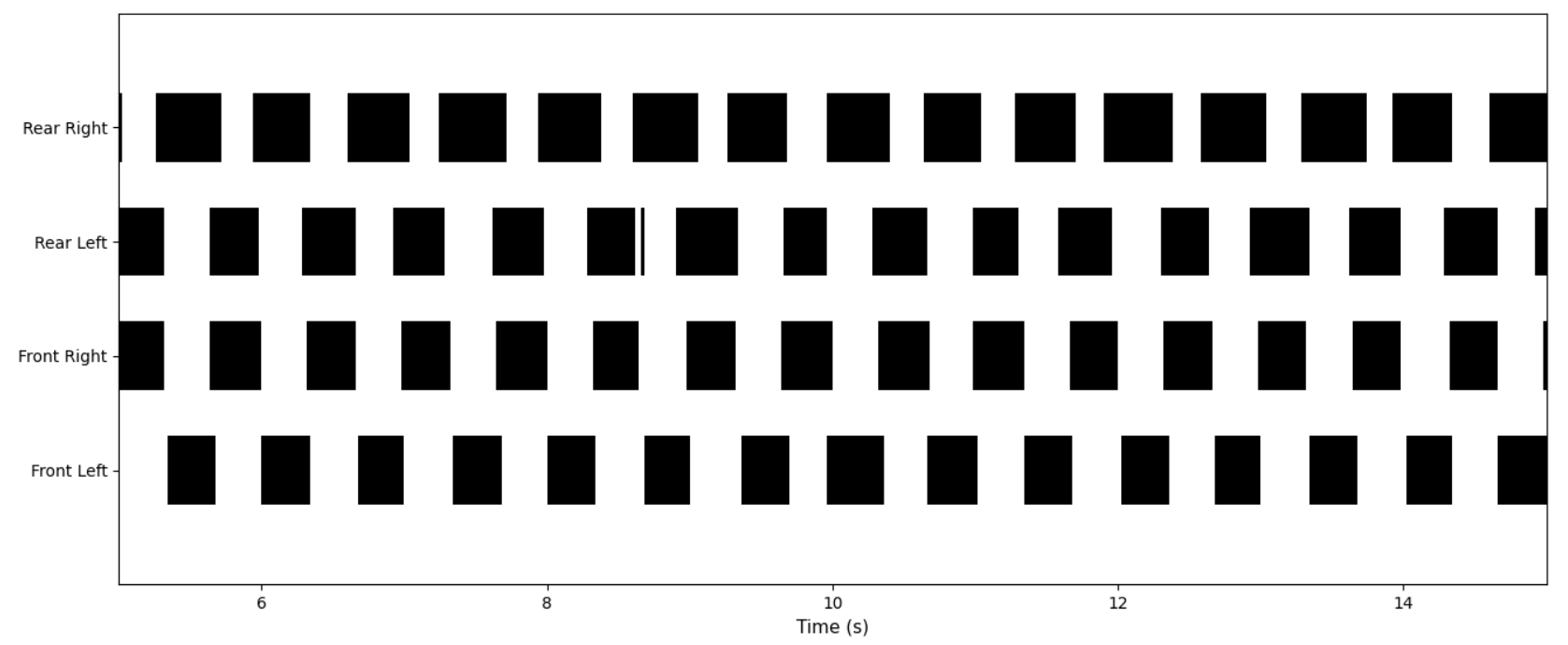}
    \caption{Go1(ZS) Foot contact sequence on Flat terrain- black regions are periods of foot contact}
    \label{fig:image1}
\end{figure}

\begin{figure}[t]
    \centering
    \includegraphics[width=0.8\linewidth]{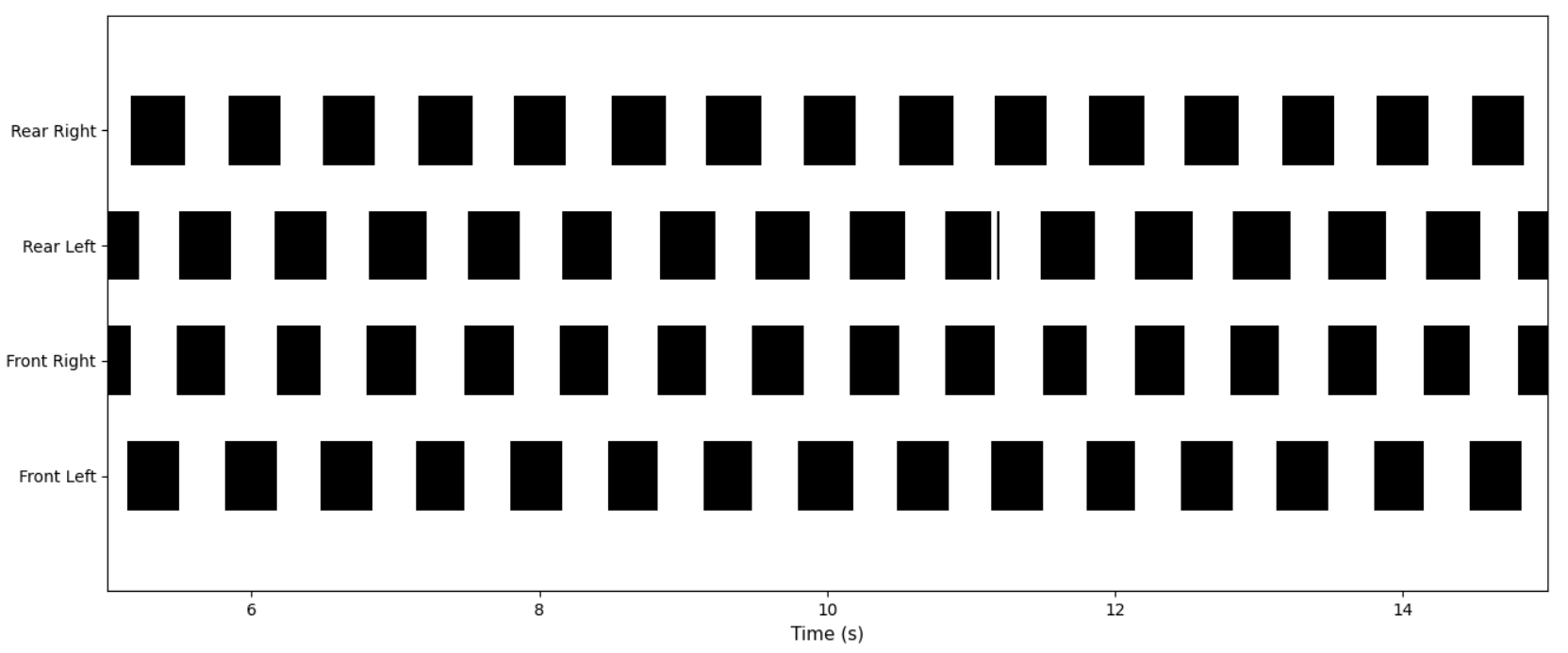}
    \caption{Stoch3(SO) Foot contact sequence on Flat terrain- black regions are periods of foot contact}
    \label{fig:image2}
\end{figure}

\paragraph{Experiment Settings.}
We consider five distinct data distribution settings for training:

\begin{itemize}
    \item \textbf{Setting 1:} Flat-terrain data from B2; stair-climbing data from Aliengo and Stoch3.
    \item \textbf{Setting 2:} Flat-terrain data from Go1 and Stoch3; stair-climbing data from B2.
    \item \textbf{Setting 3:} Flat-terrain data from Go1; stair-climbing data from Aliengo and Stoch3.
    \item \textbf{Setting 4:} Flat-terrain data from Stoch3; stair-climbing data from Go1.
    \item \textbf{Setting 5}: Both flat terrain and stair-climbing data from Aliengo, B2, and Stoch3.

\end{itemize}

\paragraph{Stair Climbing Generalization Analysis.}

We draw four key insights from the results in Table \ref{tab:stair_results}, supported by base trajectories and gait visualizations across robots.

\begin{enumerate}
    \item \textbf{Full Diversity Training Enables Strong Zero-Shot Transfer.}  
    Setting~5, which includes flat and stair data from three diverse robots, yields the best zero-shot (ZS) generalization to unseen embodiments and out-of-distribution (OOD) stairs. For example, Go1 in Setting~5 climbs 21\,cm stairs ZS, whereas it fails at the same height in Setting~2 as depicted in the Figure \ref{fig:photo1}. Compared to Setting~4, where all ZS robots fail beyond 17\,cm, Setting~5 shows clear cross-embodiment transfer.

    \item \textbf{Stair-Trained Policies Generalize Beyond Training Range.}  
    Stair-only (SO) robots generalize to OOD stairs beyond their training limit of 17\,cm. In Setting~5, B2 and Stoch3 successfully traverse 25\,cm and 29\,cm stairs, showing strong terrain extrapolation.

    \item \textbf{Zero-Shot Robots Can Outperform Stair Specialists.}  
    In some cases, ZS robots surpass SO-trained ones. For instance, in Setting~3, B1 (ZS) climbs 29\,cm stairs, while Aliengo (SO) fails. Figure \ref{fig:photo2} suggests more adaptive motions in ZS policies due to morphology-driven robustness.

    \item \textbf{Flat-Only Policies Exhibit Stair Climbing Generalization.}  
    Flat-only (FO) policies show generalization to stairs. In Setting~2, Stoch3 (FO) climbs 25\,cm OOD stairs despite no elevation exposure during training, indicating transferable skills like stable gait and foot placement.

    \item \textbf{Emergence of Cyclic Gaits Across Robots.}  
    Gait plots (Figures \ref{fig:image1}, \ref{fig:image2}) show structured, cyclic patterns in both ZS (Go1) and SO (Stoch3) policies across terrains.
\end{enumerate}

\begin{figure}[t]
    \centering
    \begin{subfigure}[b]{0.48\linewidth}
        \centering
        \begin{tabular}{|c|ccc||ccc|}
            \hline
            \textbf{Robot} & \multicolumn{3}{c||}{\textbf{Smooth Slopes}} & \multicolumn{3}{c|}{\textbf{Rough Slopes}} \\
            \cline{2-7}
            & \textbf{$25^\circ$} & \textbf{$30^\circ$} & \textbf{$40^\circ$} & \textbf{$25^\circ$} & \textbf{$30^\circ$} & \textbf{$40^\circ$} \\
            \hline
            Go1     & \cmark & \xmark & \xmark & \cmark & \xmark & \xmark \\
            Stoch5  & \cmark & \cmark & \cmark & \cmark & \cmark & \cmark \\
            Aliengo & \cmark & \cmark & \cmark & \cmark & \cmark & \cmark \\
            Stoch3  & \cmark & \cmark & \xmark & \cmark & \cmark & \xmark \\
            B2      & \cmark & \cmark & \cmark & \cmark & \cmark & \cmark \\
            B1      & \cmark & \cmark & \cmark & \cmark & \cmark & \cmark \\
            \hline
        \end{tabular}
        \caption{Zero-shot slope traversal results.}
        \label{tab:slope_eval}
    \end{subfigure}
    \hfill
    \begin{subfigure}[b]{0.48\linewidth}
        \centering
        \includegraphics[width=\linewidth]{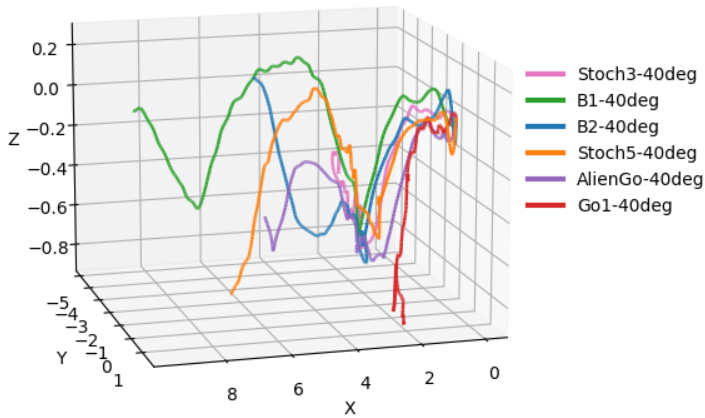}
        \caption{Base trajectory on $40^\circ$ rough slope.}
        \label{fig:slope_traj}
    \end{subfigure}
    \caption{Binary performance assessment and visual evaluation of zero-shot generalization to novel slopes.}
    \label{fig:slope_combo}
\end{figure}

\subsection{Generalization to Novel Terrains}

We evaluate zero-shot generalization of our locomotion policies to smooth and rough inclined slopes at angles of $25^\circ$, $30^\circ$, and $40^\circ$, none of which appeared in training. All robots are deployed in a fully zero-shot setting (\textit{unseen robot}~$\times$~\textit{terrain}). Table in Figure~\ref{tab:slope_eval} summarizes binary success (\cmark) or failure (\xmark) on each slope.

  \textbf{Consistent High-Performers.}  
    Stoch5, Aliengo, B2, and B1 succeed on all smooth and rough slopes up to $40^\circ$, demonstrating exceptionally robust zero-shot slope traversal across morphologies and terrain irregularities.

   \textbf{Minimal Impact of Roughness for Robust Policies.}  
    For high-performing robots, undulating terrain (see  Fig. \ref{fig:slope_traj}) does not degrade performance compared to smooth slopes. This suggests that our policy architecture—and particularly the cross-robot training—captures slope-invariant locomotion strategies.

   \textbf{Zero-Shot Emergence of Adaptive Behavior.}  
Despite no slope data during training, policies exhibit adaptive base movement and posture control on inclines. Base trajectory plots (Figure~\ref{fig:slope_traj}) reveal smooth and progressive elevation changes, indicating stable and coordinated climbing behavior in a zero-shot setting.

\textbf{Hardware deployment.} We deployed a cloned policy on the Unitree Go1 and Stoch5. Go1 policy (Setting 5 in Table \ref{tab:stair_results}) demonstrated a zero-shot transfer for both flat-ground and staircases (15cm height). Stoch 5 policy demonstrated robust walking on flat-ground and slopes. Detailed results will be provided in the supplementary material.

\section{Conclusion and Limitations}
\label{sec:limitation}

Currently, GRoQ-LoCO is focused on robots with comparable kinematic setups. Extending the framework to quadrupeds with more diverse morphologies — such as those with different leg proportions or joint arrangements — remains an exciting direction for future work. Furthermore, while our current system is based on proprioceptive feedback, integrating exteroceptive inputs like vision could allow the policy to become more visually aware, enhancing its ability to navigate complex environments and adapt to dynamic terrain. 
Another area for exploration involves expanding the approach beyond quadruped robots to other types of legged robots, such as hexapods or bipeds, where the dynamics of locomotion may present new challenges. Addressing these aspects will further enhance the versatility and generalization capabilities of our locomotion policies.

\acknowledgments{This research is funded by AI \& Robotics Technology Park (ARTPARK), India}




\bibliography{example}


\appendix
\section*{Appendix}
\addcontentsline{toc}{section}{Appendix}



\section{Extended Results section}
We report additional zero-shot results beyond those presented in the main paper.
The physical parameters of all quadruped robots, including those used in the main paper, are provided in Table~\ref{tab:robot_params}.
Table~\ref{tab:supplementary_stair_results_lite3_x30} reports zero-shot results on stair environments using two new quadruped robots: Lite3 and X30.
Table~\ref{tab:slope_eval_new_robots} further shows zero-shot performance on slope terrains (smooth and rough) across robots.

All experiments are conducted with a commanded forward velocity of $\mathbf{1\,\mathrm{m/s}}$ along the $x$-axis.

\begin{table}[ht]
\centering
\scriptsize
\begin{tabular}{|l|c|c|c|c|c|c|c|c|c|}
\hline
\textbf{Parameter} 
& \textbf{A1} & \textbf{Go1} & \textbf{Aliengo} & \textbf{Stoch3} & \textbf{B1} & \textbf{B2}
& \textbf{Stoch5} & \textbf{Lite} & \textbf{X30} \\
\hline
Total weight (kg)         & 12  & 13  & 21  & 25 & 50 & 60  & 70   & 13  & 56   \\
Base length (m)           & 0.40 & 0.38 & 0.65 & 0.54 & 0.92 & 0.80 & 0.67 & 0.53 & 0.90 \\
Base width (m)            & 0.19 & 0.16 & 0.15 & 0.20 & 0.24 & 0.24 & 0.26 & 0.20 & 0.30 \\
Height, fully standing (m)& 0.40 & 0.40 & 0.48 & 0.50 & 0.63 & 0.64 & 0.55 & 0.40 & 0.47 \\
Thigh Length (m)          & 0.20 & 0.22 & 0.26 & 0.30 & 0.35 & 0.35 & 0.35 & 0.2 & 0.3 \\
Calf Length (m)           & 0.20 & 0.22 & 0.26 & 0.35 & 0.35 & 0.35 & 0.35 & 0.21 & 0.31 \\
\hline
\end{tabular}
\vspace{10pt}
\caption{Comparison of quadruped robot parameters}
\label{tab:robot_params}
\end{table}

\begin{table}[ht]
\centering
\begin{tabular}{|c|c|c|c|c|c|c|c|}
\hline
\textbf{Setting} & \textbf{Robot} & \textbf{Mode} & \textbf{13 cm} & \textbf{17 cm} & \textbf{21 cm (OOD)} & \textbf{25 cm (OOD)} & \textbf{29 cm (OOD)} \\
\hline
\multirow{2}{*}{1} & Lite3 & ZS & \cmark & \cmark & \xmark & \xmark & \xmark \\ 
& X30 & ZS & \cmark & \cmark & \cmark & \cmark & \xmark \\ 
\hline
\multirow{2}{*}{2} & Lite3 & ZS & \xmark & \xmark & \xmark & \xmark & \xmark \\ 
& X30 & ZS & \cmark & \cmark & \cmark & \xmark & \xmark \\ 
\hline
\multirow{2}{*}{3} & Lite3 & ZS & \cmark & \xmark & \xmark & \xmark & \xmark \\ 
& X30 & ZS & \cmark & \cmark & \cmark & \xmark & \xmark \\ 
\hline
\multirow{2}{*}{4} & Lite3 & ZS & \cmark & \cmark & \cmark & \xmark & \xmark \\ 
& X30 & ZS & \cmark & \cmark & \xmark & \xmark & \xmark \\ 
\hline
\multirow{2}{*}{5} & Lite3 & ZS & \cmark & \cmark & \xmark & \xmark & \xmark \\ 
& X30 & ZS & \cmark & \cmark & \cmark & \cmark & \xmark \\ 
\hline
\end{tabular}
\vspace{3 mm} 
\caption{Zero-Shot evaluation on stair environments for Lite3 and X30 robots.
}
\label{tab:supplementary_stair_results_lite3_x30}
\end{table}

\begin{table}[ht]
    \centering
    \begin{tabular}{|c|ccc||ccc|}
        \hline
        \textbf{Robot} & \multicolumn{3}{c||}{\textbf{Smooth Slopes}} & \multicolumn{3}{c|}{\textbf{Rough Slopes}} \\
        \cline{2-7}
        & \textbf{25°} & \textbf{30°} & \textbf{40°} & \textbf{25°} & \textbf{30°} & \textbf{40°} \\
        \hline
        Lite3     & \cmark & \cmark & \xmark & \cmark & \cmark & \xmark \\
        X30  & \cmark & \cmark & \cmark & \cmark & \cmark & \cmark \\
        \hline
    \end{tabular}
    \vspace{10pt}
    \caption{Zero-shot slope traversal results.}
    \label{tab:slope_eval_new_robots}
\end{table}


\section{Robot Encoding vs No Robot Encoding}
The main paper presents five cross-robot training settings involving various combinations of flat-terrain and stair-climbing data. We provide extended analysis for \textbf{Setting 1} (flat data from B2; stair data from Aliengo and Stoch3), comparing models trained with explicit robot encodings under otherwise identical conditions. Table~\ref{tab:stair_results_robo_enc} shows the stair-climbing results using policies trained with explicit robot encodings.

We use robot encodings as part of observation input to the model, derived from predefined metadata as shown in the table \ref{tab:robot_params}. However, we observe that explicit robot encoding does not lead to reliable generalization. Specifically, robots like \textbf{Go1} and \textbf{Lite3}, whose embeddings are significantly different from those seen during training, fail to exhibit meaningful behavior. These robots attempt to walk but immediately collapse with erratic actions, indicating poor transfer to out-of-distribution embeddings.

In contrast, we observe some degree of skill fusion in the \textbf{FO (Flat-only)} robots. For instance, \textbf{B2}, trained solely on flat terrain, can climb stairs up to 21\,cm. Robots with similar embeddings—\textbf{B1}, \textbf{Stoch5}, and \textbf{X30}—also demonstrate similar stair-climbing behavior in zero-shot settings. This suggests that proximity in the embedding space can enable generalization.

\begin{figure}[ht]
    \centering
    \includegraphics[width=0.8\linewidth]{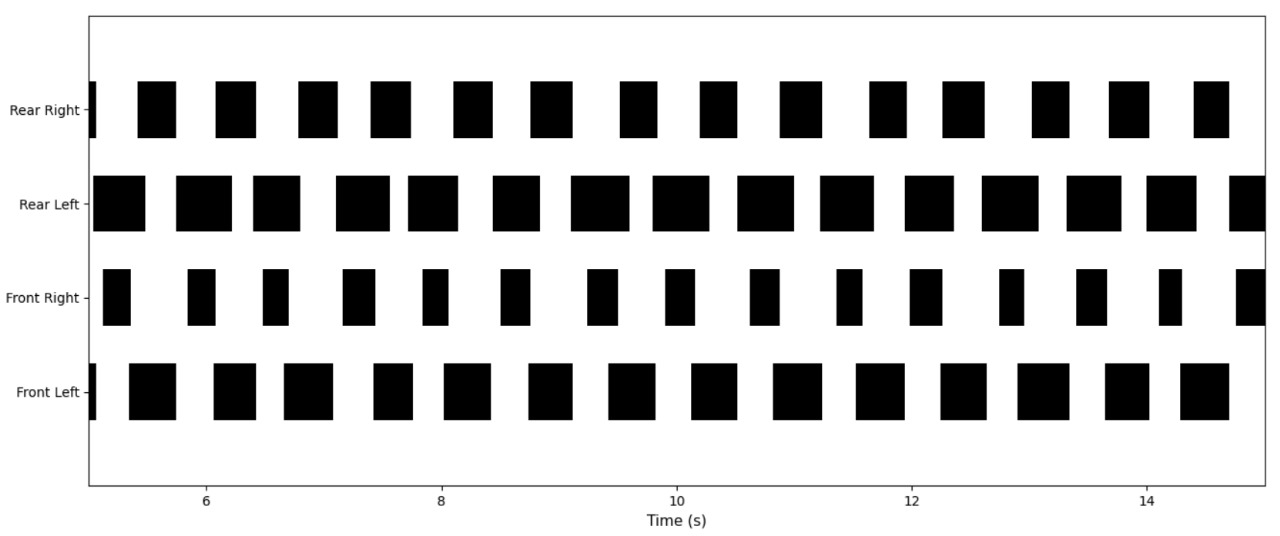}
    \caption{Stoch5(ZO) Foot contact sequence on Flat terrain - No Robot Encoding}
    \label{fig:mule_no_robo_enc}
\end{figure}

\begin{figure}[ht]
    \centering
    \includegraphics[width=0.8\linewidth]{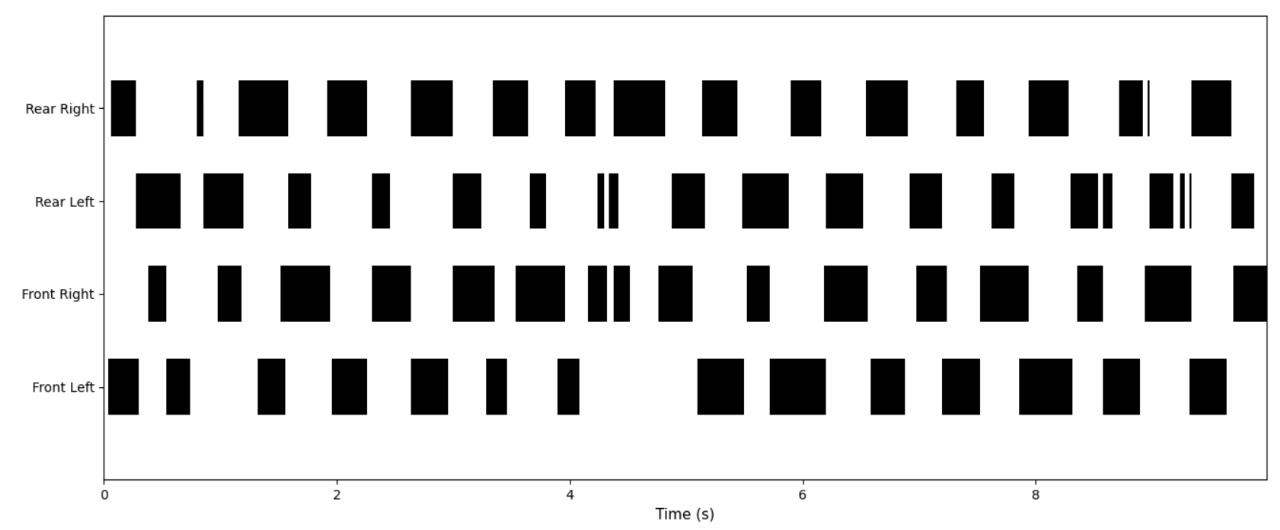}
    \caption{Stoch5(ZO) Foot contact sequence on Flat terrain - Robot Encoding}
    \label{fig:mule_robo_enc}
\end{figure}

\textbf{Gait Analysis:} We observe clear periodic gaits for \textbf{ZS (Zero-shot)} robots (Stoch5) in the no-robot-encoding case (Figure \ref{fig:mule_no_robo_enc}), indicating successful skill fusion. In contrast, the robot-encoding variant (Figure \ref{fig:mule_robo_enc}) shows disrupted periodicity, suggesting weaker generalization.

\textbf{Hypothesis:} We hypothesize that more diverse training data spanning a broader range of robot morphologies could improve the learned embedding space. Additionally, future work could explore varying the embedding dimensionality or structure to enhance generalization and avoid overfitting to specific robot identities.

\begin{table}[t]
    \centering
\begin{tabular}{|c|c|c|c|c|c|c|c|}
\hline
\textbf{Setting} & \textbf{Robot} & \textbf{Mode} & \textbf{13 cm} & \textbf{17 cm} & \textbf{21 cm (OOD)} & \textbf{25 cm (OOD)} & \textbf{29 cm (OOD)} \\
\hline
  & Go1     & ZS & \xmark & \xmark & \xmark & \xmark & \xmark \\
  & Stoch5  & ZS & \cmark & \cmark & \cmark & \xmark & \xmark \\
 & B1      & ZS & \cmark & \cmark & \cmark & \xmark & \xmark \\
 1 & B2      & FO & \cmark & \cmark & \cmark & \xmark & \xmark \\
  & Aliengo & SO & \cmark & \cmark & \cmark & \xmark & \xmark \\
  & Stoch3  & SO & \cmark & \cmark & \cmark & \xmark & \xmark \\
  & Lite3  & ZS & \xmark & \xmark & \xmark & \xmark & \xmark \\
  & X30  & ZS & \cmark & \cmark & \cmark & \xmark & \xmark \\
\hline
\end{tabular}
\vspace{3 mm}  

\caption{Evaluation on stair environments with increasing difficulty (13–29 cm step heights). A checkmark (\cmark) indicates successful climbing of 8 stairs, and a cross (\xmark) indicates failure.}
\label{tab:stair_results_robo_enc}
\end{table}

\section{Adaptive Loss and Residual Variance Dynamics}

To analyze the effectiveness of the adaptive loss in modeling complex locomotion behaviors across diverse terrains, we investigate the learned per-joint residual distributions after $3000$ training steps. Our loss function dynamically adjusts the contribution of each joint prediction by learning a joint-specific variance $\sigma^2$, which controls the weighting of the corresponding residuals during optimization.

\subsection*{Per-Joint Residual Distribution Analysis}

The residuals approximately follow zero-mean Gaussian distributions, with their learned variances reflecting the relative modeling difficulty of each joint.

We visualize representative residual distributions for selected joints after \textbf{$3000$} training steps under the adaptive loss framework. The plots show per-joint residuals fitted to zero-mean Gaussians, with variance values learned through joint-wise adaptive weighting. The selected joints span different joint types (abdomen, hip, knee) and leg locations (front/back, left/right), ensuring a balanced representation across the body.

\subsection*{Visualization}

\begin{figure*}[t]
  \centering
  \begin{tabular}{ccc}
    \subfloat[0 steps (All joints)]{\includegraphics[width=0.3\textwidth]{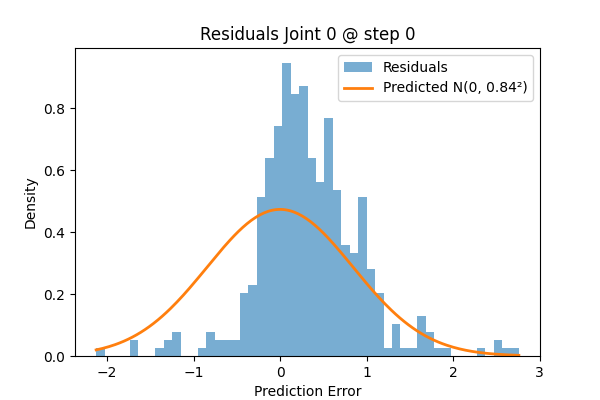}} &
    \subfloat[FL Abd]{\includegraphics[width=0.3\textwidth]{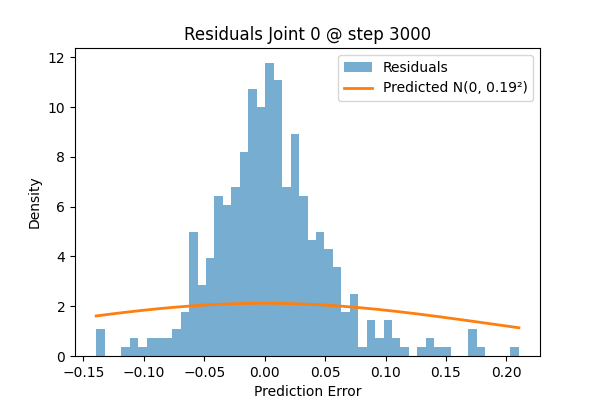}} &
    \subfloat[FL Knee]{\includegraphics[width=0.3\textwidth]{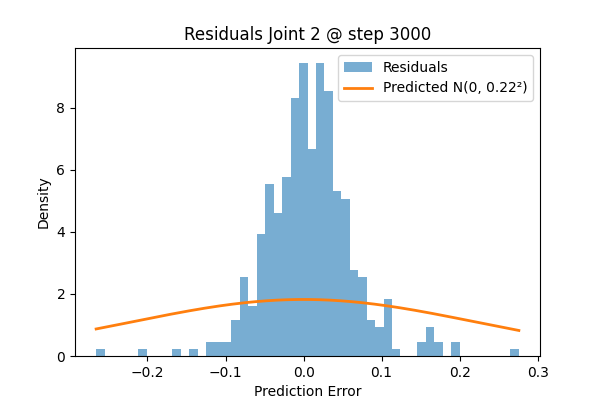}}
 \\
    \subfloat[FR Hip]{\includegraphics[width=0.3\textwidth]{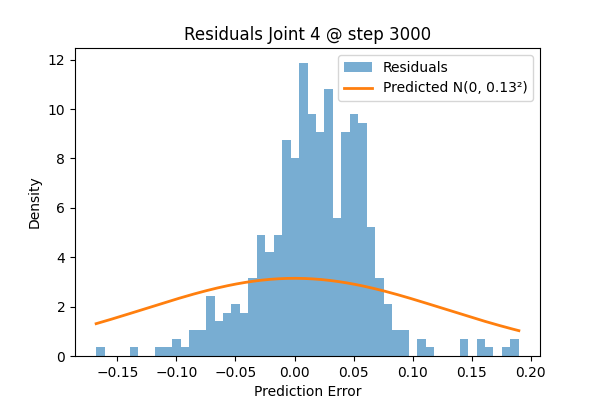}} &
    \subfloat[BL Knee]{\includegraphics[width=0.3\textwidth]{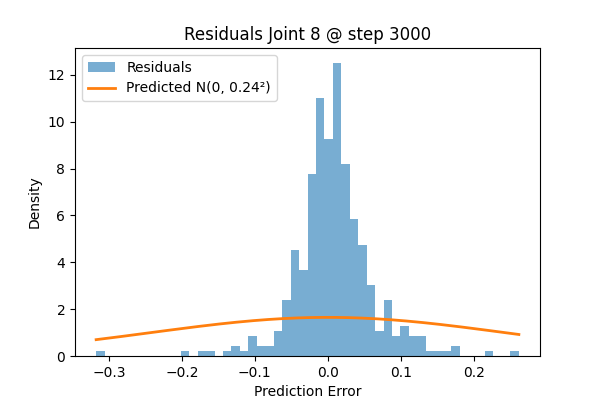}} &
    \subfloat[BR Abd]{\includegraphics[width=0.3\textwidth]{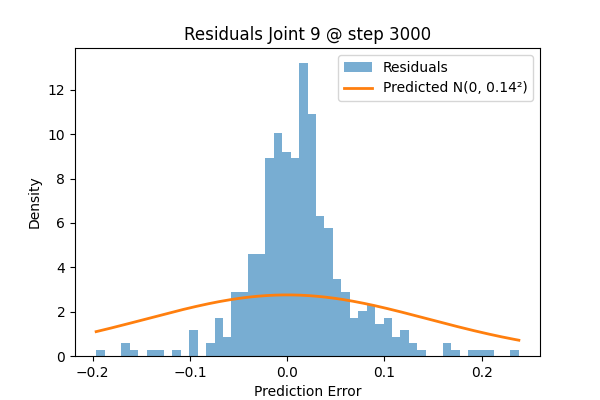}} \\
  \end{tabular}
\caption{
Kernel density plots of residual errors across representative joints. 
Subfigure (a) shows the initial high-variance residual distribution across all joints at step 0, 
while (b–f) show joint-specific residuals after 3000 training steps, revealing progressive variance adaptation.
}
  \label{fig:residual_table}
\end{figure*}

Figure~\ref{fig:residual_table} shows kernel density plots of the residuals after 3000 training iterations. The shift in density and narrowing of variance highlights the transition from cautious early-stage learning to confident, fine-tuned predictions in later training. These plots further validate the utility of per-joint adaptive weighting for both stable and high-variance locomotion regimes.

\section{Architectural Details}

\subsection{Model Architecture}
Table \ref{tab:hyperparam_tab} displays the hyperparameters of the model.
Our architecture follows a modular and interpretable design intended for generalist quadruped locomotion. The architecture is composed of the following blocks:

\begin{itemize}
    \item \textbf{Observation Encoder:} A linear layer with ELU activation followed by LayerNorm that encodes raw observations and estimated velocity into an embedding space of dimension $\texttt{emb\_dim} = 128$.
    \item \textbf{Observation Attention Block:} Multi-head self-attention with 4 heads is applied over a temporal window of past observations using fixed sinusoidal positional embeddings.
    \item \textbf{GRU Block:} A GRU module with input size $2 \times \texttt{emb\_dim}$ and hidden size $\texttt{emb\_dim}$ encodes the temporal evolution of attended observation features.
    \item \textbf{GRU Attention Block:} Similar to the observation attention block, this layer captures temporal dependencies over the GRU outputs.
    \item \textbf{MLP Head:} A 3-layer MLP maps the concatenated context vector (raw observation encoding, GRU output, and GRU attention output) into action space.
\end{itemize}

A schematic diagram of the model structure is given in the main paper.

\vspace{1em}

\begin{table}[t]
\centering
\begin{tabular}{lcc}
\toprule
\textbf{Component} & \textbf{Value} & \textbf{Remarks} \\
\midrule
Embedding Dimension ($\texttt{emb\_dim}$) & 64 & Used in the observation encoder module \\
Observation Dimensionality & 54 + 3 & Observation + commanded velocity \\
GRU Hidden Size & 64 & Matching the embedding dim \\
Attention Heads & 4 & In both attention blocks \\
Attention Window Size & 100 & For both obs and GRU attention \\
MLP Hidden Size & 256 & 2-layer ELU MLP head \\
Optimizer & Adam & Standard \\
Learning Rate & $1\mathrm{e}{-3}$ & Fixed across experiments \\
Batch Size & 400 &  Per forward pass (batch of 400 trajectories) \\
\bottomrule
\bottomrule
\end{tabular}
\vspace{10pt}
\caption{Model Hyperparameters}
\label{tab:hyperparam_tab}

\end{table}

\vspace{1em}
\subsection{Attention Design: Single Query Approach}

In our model, two types of history are maintained: the observation history and the GRU history. These histories allow the model to capture temporal dependencies across observations and GRU outputs, respectively.

\textbf{Observation History:} The observation history consists of the most recent observations, stored over a window of size \(W\) = 100. For each timestep, the encoded observation is appended to the history. When the window exceeds its size, the earliest observation is removed to make room for the new one.

Let the observation history be denoted as:

\[
\mathbf{o} = [o_1, o_2, \dots, o_W]
\]

Where \(o_t\) represents the encoded observation at timestep \(t\), and \(W\) is the maximum number of timesteps the history holds.

\textbf{GRU History:} Similarly, the GRU history stores the output of the GRU at each timestep. The GRU processes the concatenated observation and its associated attention context, and the resulting output is stored in a history buffer of size \(W\) = 100.

Let the GRU history be denoted as:

\[
\mathbf{h} = [h_1, h_2, \dots, h_W]
\]

Where \(h_t\) represents the GRU output at timestep \(t\), and \(W\) is the size of the GRU history window.

\textbf{Single Query Attention:} After storing these histories, the attention mechanism uses the most recent timestep from the observation history and GRU history for the query. This reduces computational complexity by using only the latest timestep for attention calculations, rather than the entire history.

Let \(q_o = o_W\) and \(q_h = h_W\) be the queries from the observation and GRU histories, respectively. The attention computation is performed as follows:

\[
\text{Attention}(q_o, \mathbf{o}) = \text{Softmax}\left( \frac{q_o K_o^T}{\sqrt{d_k}} \right) V_o
\]

\[
\text{Attention}(q_h, \mathbf{h}) = \text{Softmax}\left( \frac{q_h K_h^T}{\sqrt{d_k}} \right) V_h
\]

Where:
- \(K_o\) and \(K_h\) are the key matrices for observation and GRU histories, respectively,
- \(V_o\) and \(V_h\) are the value matrices for observation and GRU histories, and
- \(d_k\) is the dimension of the key vectors.

This single query approach is computationally lighter compared to using full attention, as it minimizes both memory and computation costs.

\vspace{1em}




\subsection{Attention Pattern Analysis Across Terrains}



\begin{figure}[ht]
\centering
\begin{tabular}{lcc}
\toprule
\textbf{Robot $\backslash$ Head} & \textbf{Head 0} & \textbf{Head 1} \\
\midrule

\textbf{Go1 (Small)} &
\includegraphics[width=0.28\textwidth]{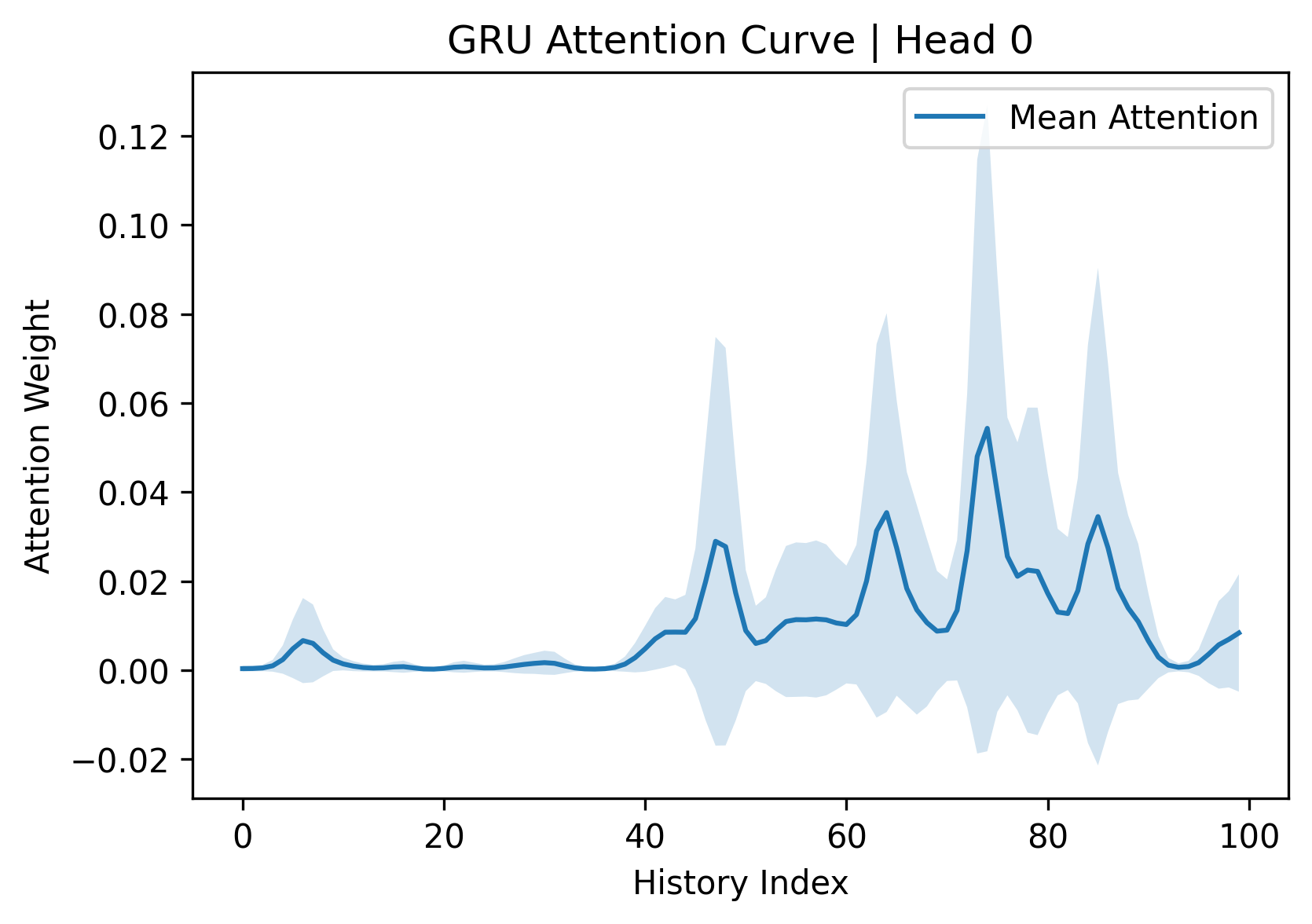} &
\includegraphics[width=0.28\textwidth]{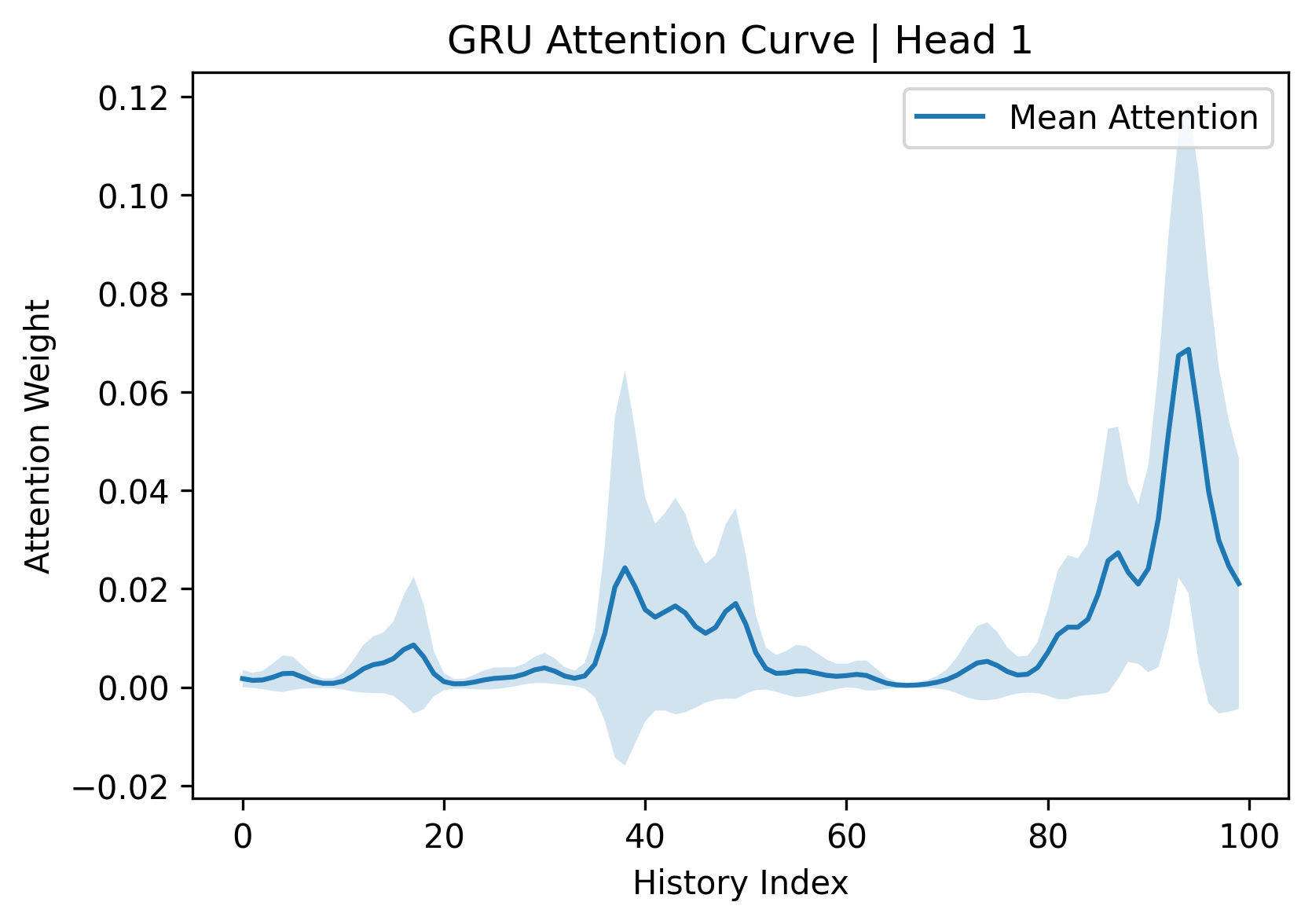} \\

\textbf{Stoch3 (Medium)} &
\includegraphics[width=0.28\textwidth]{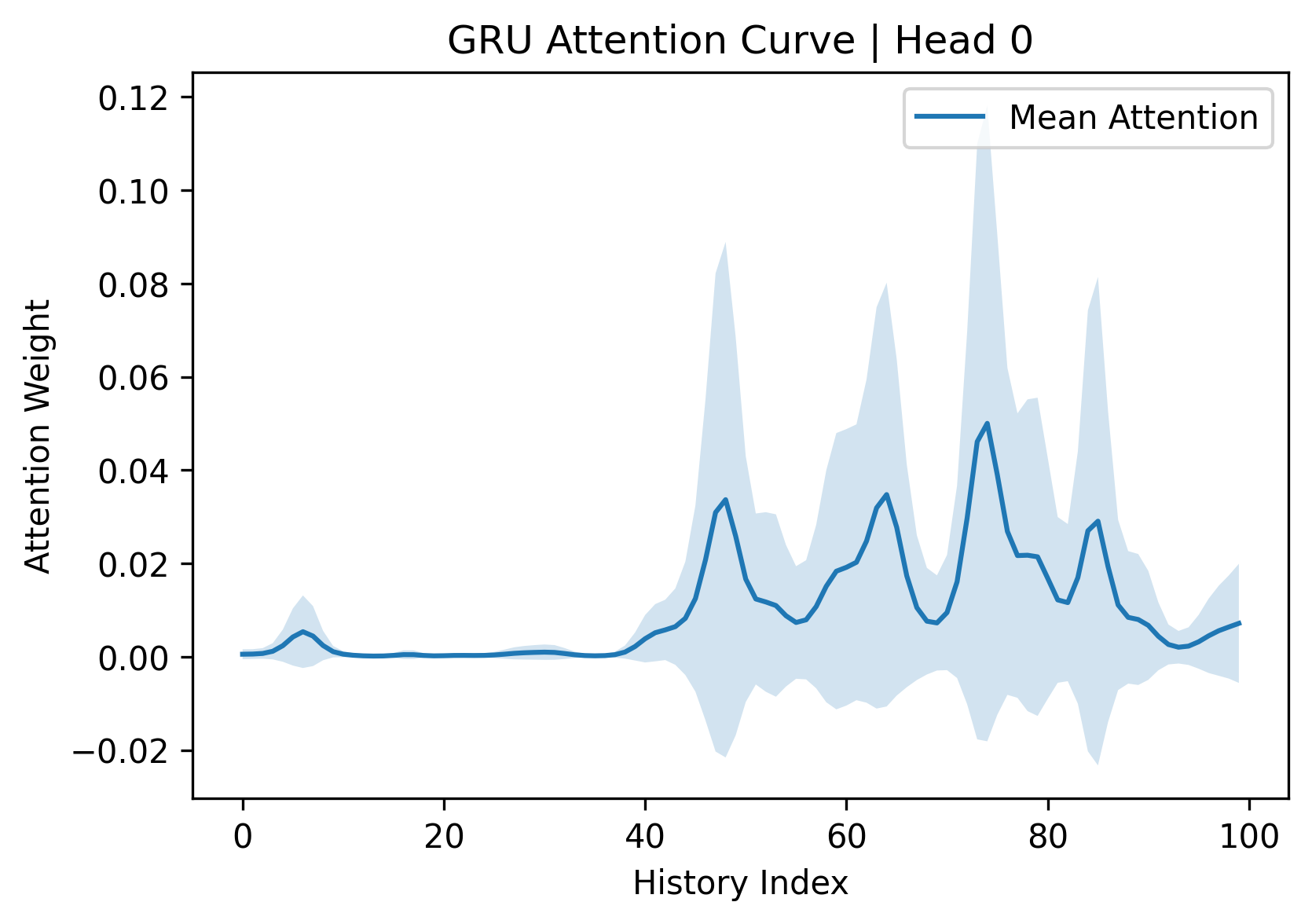} &
\includegraphics[width=0.28\textwidth]{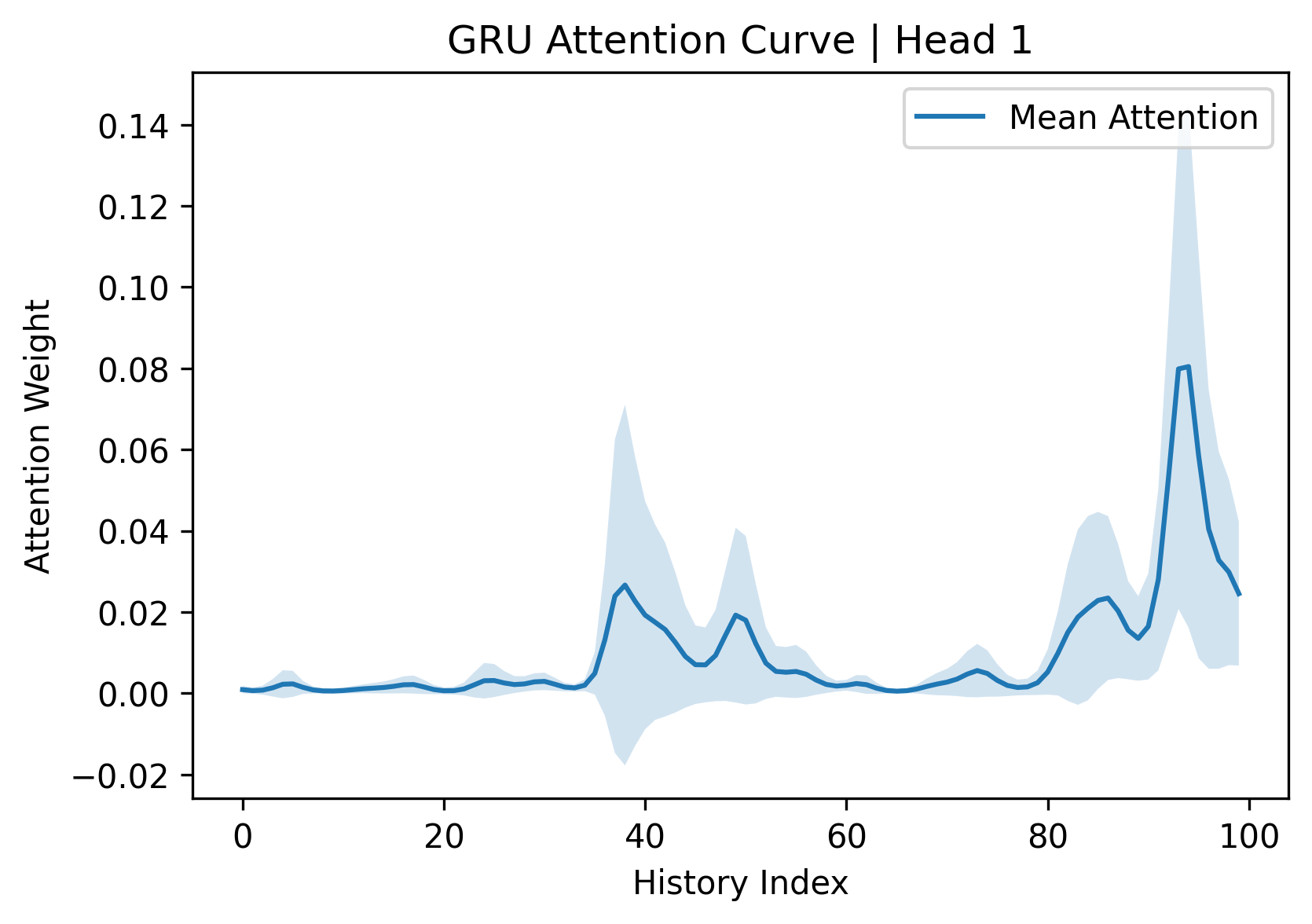} \\

\textbf{B1 (Large)} &
\includegraphics[width=0.28\textwidth]{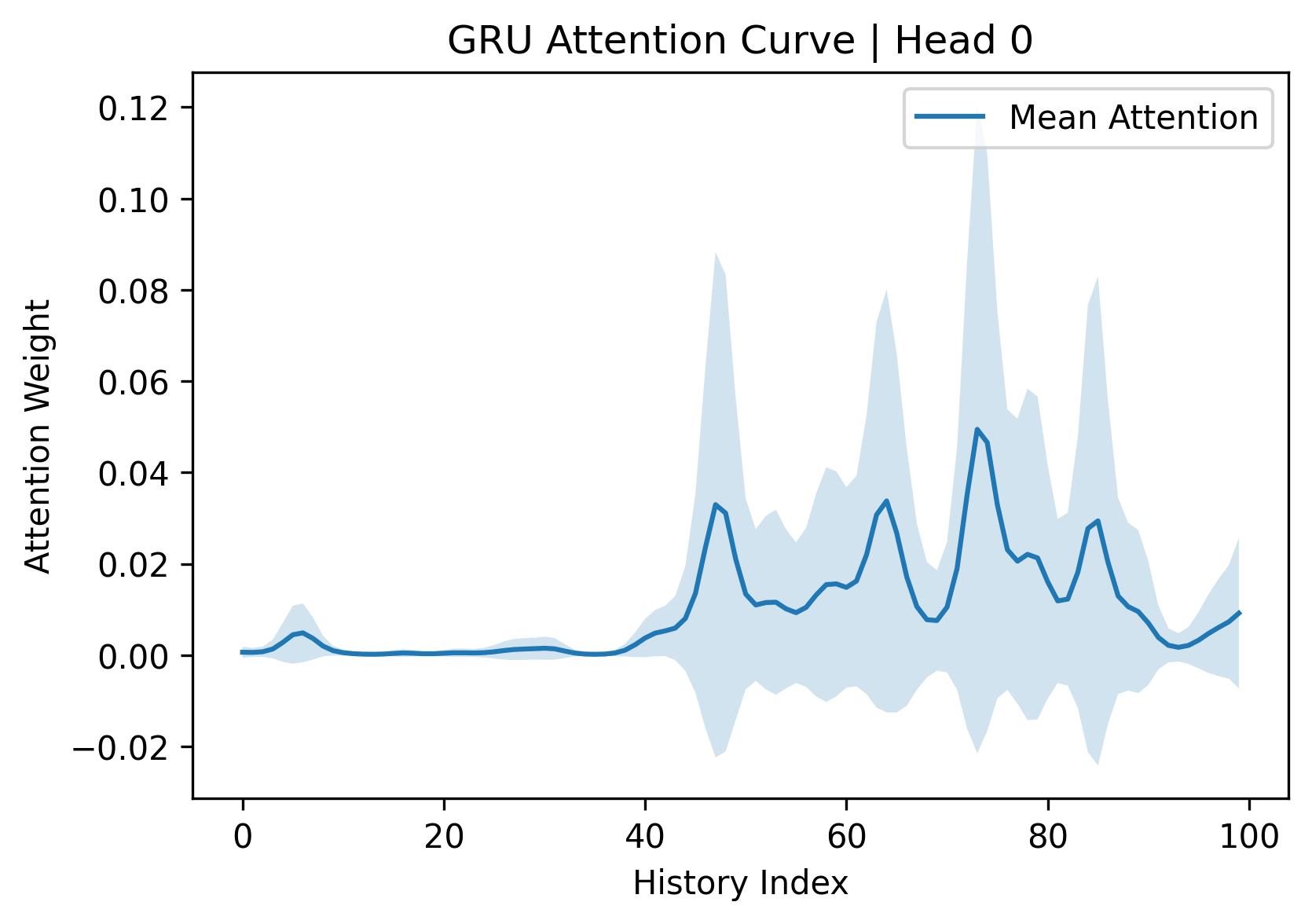} &
\includegraphics[width=0.28\textwidth]{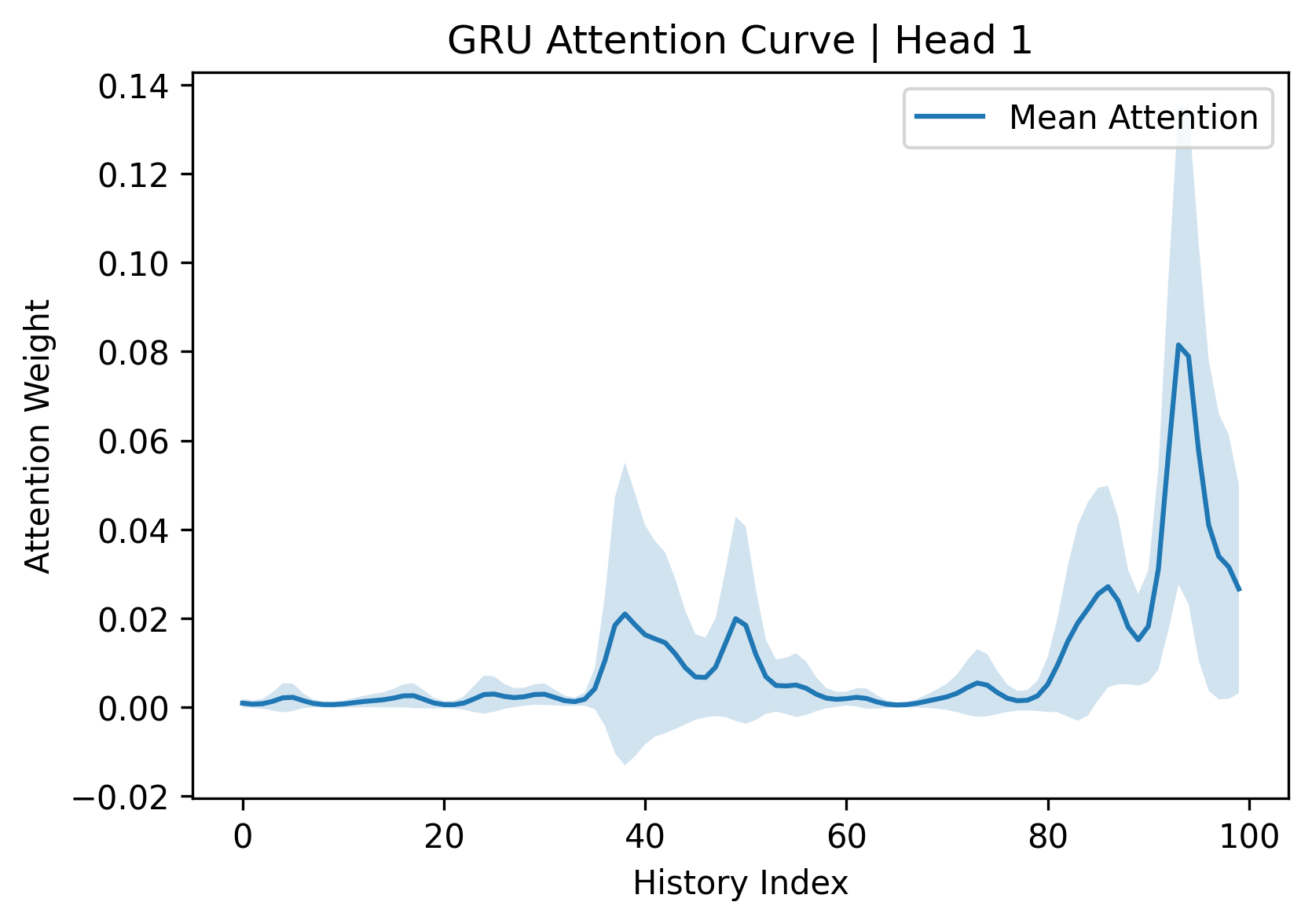} \\

\bottomrule
\end{tabular}
\vspace{3pt}
\caption{Mean GRU Attention Curves for \textbf{Flat Terrain} (17cm) Across Robots and Selected Heads}
\label{fig:flat_attention_gru}

\end{figure}
\begin{figure}[ht]
\centering

\begin{tabular}{lcc}
\toprule
\textbf{Robot $\backslash$ Head} & \textbf{Head 0} & \textbf{Head 1} \\
\midrule

\textbf{Go1 (Small)} &
\includegraphics[width=0.28\textwidth]{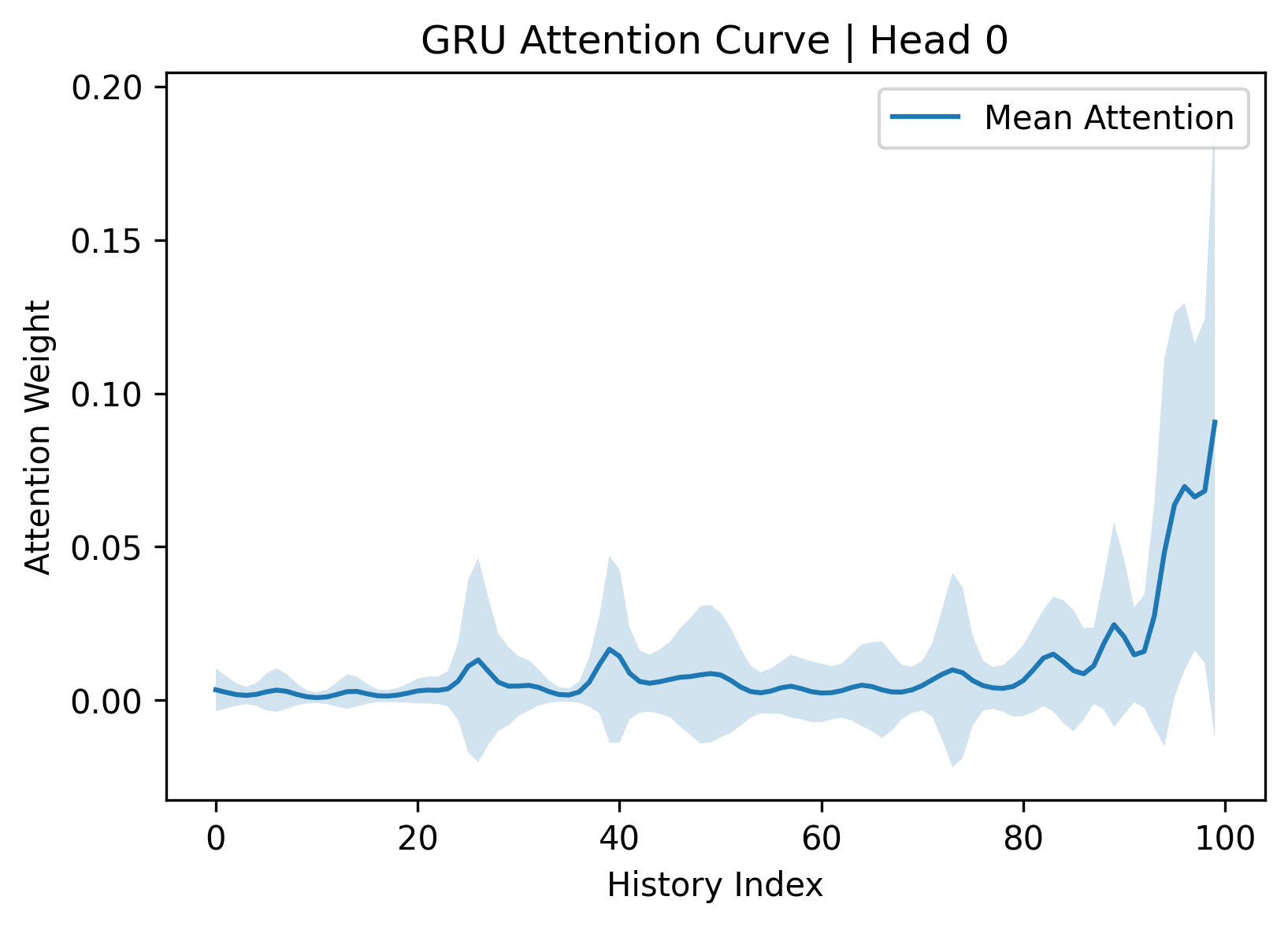} &
\includegraphics[width=0.28\textwidth]{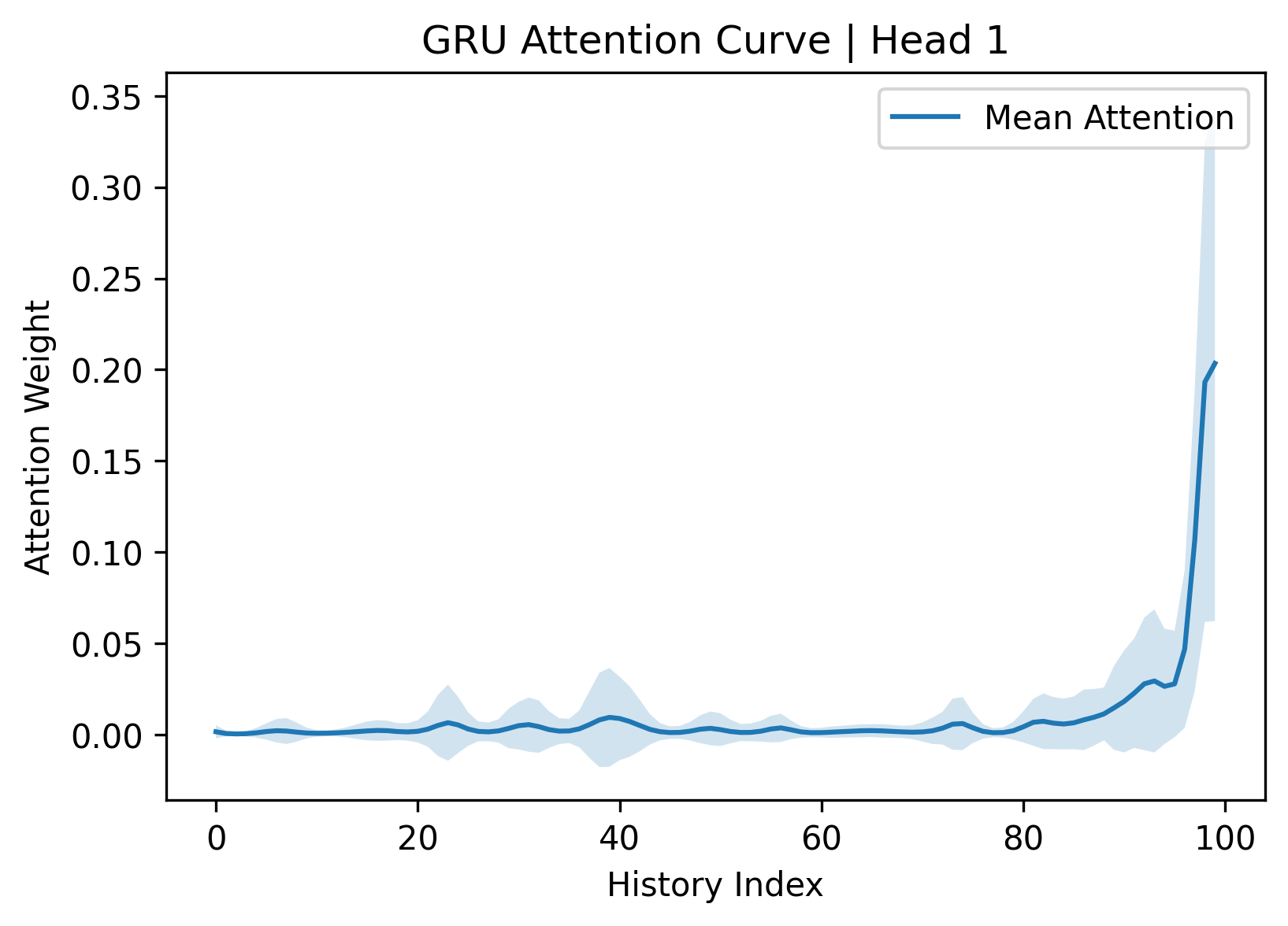} \\

\textbf{Stoch3 (Medium)} &
\includegraphics[width=0.28\textwidth]{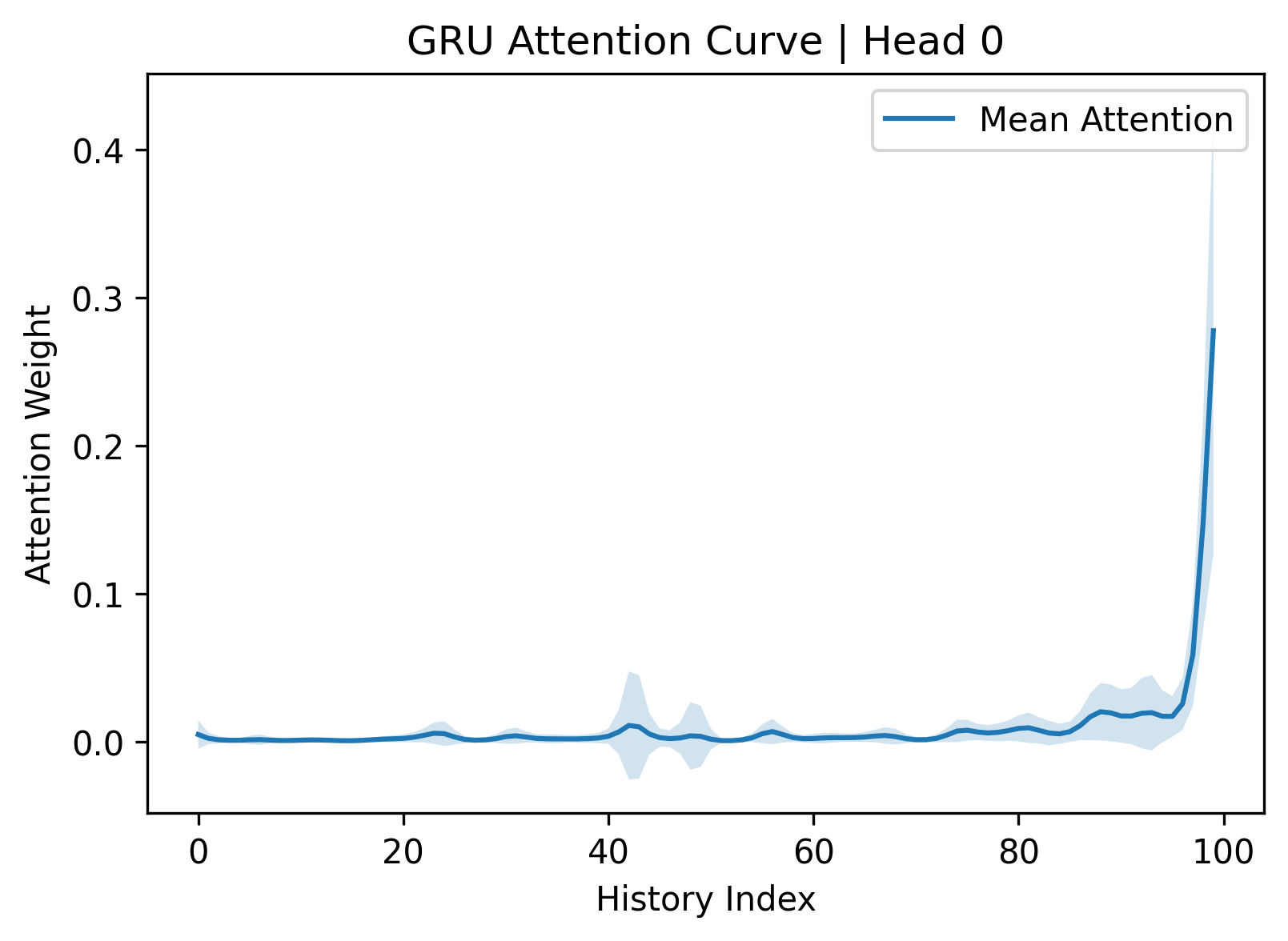} &
\includegraphics[width=0.28\textwidth]{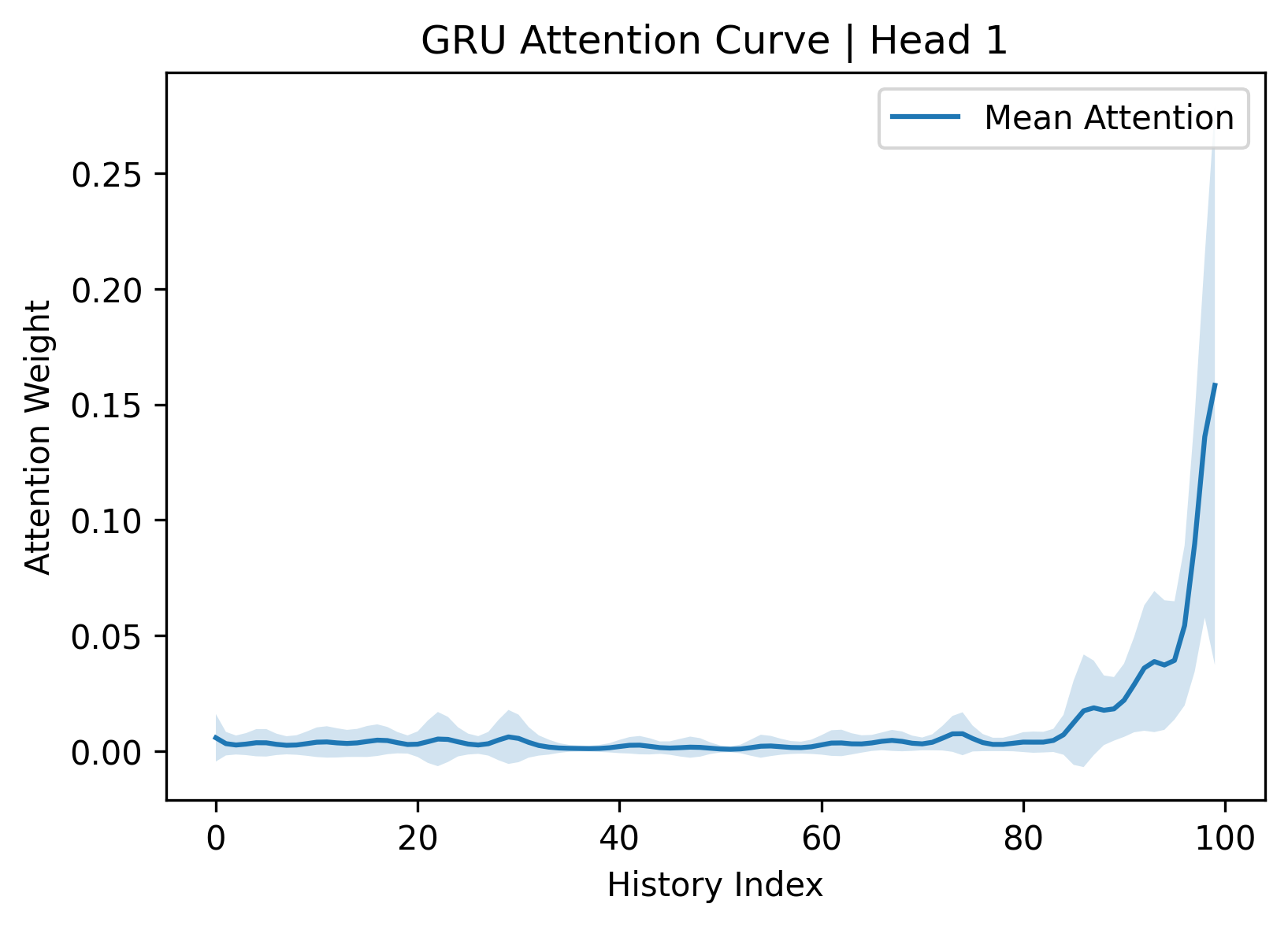} \\

\textbf{B1 (Large)} &
\includegraphics[width=0.28\textwidth]{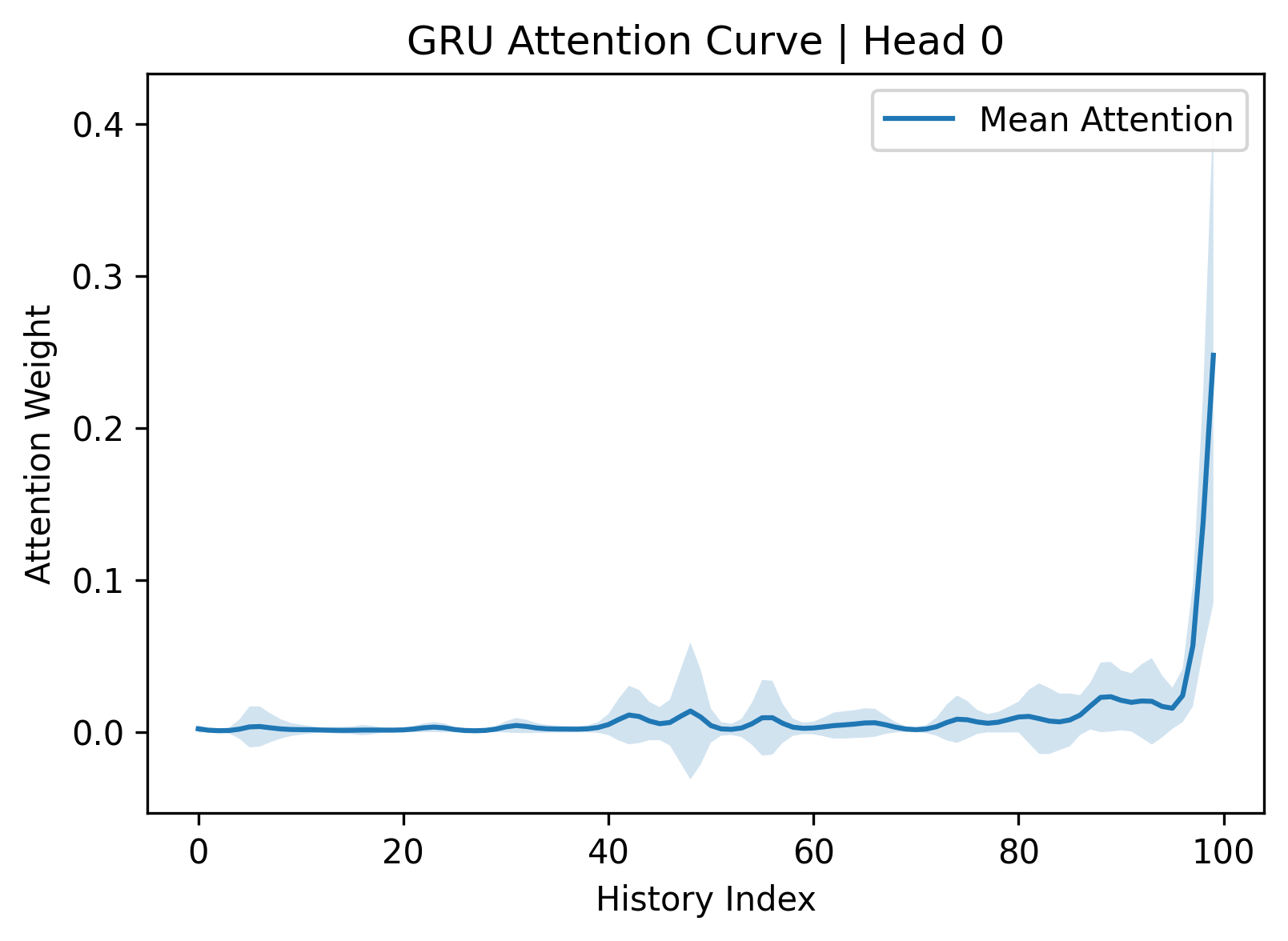} &
\includegraphics[width=0.28\textwidth]{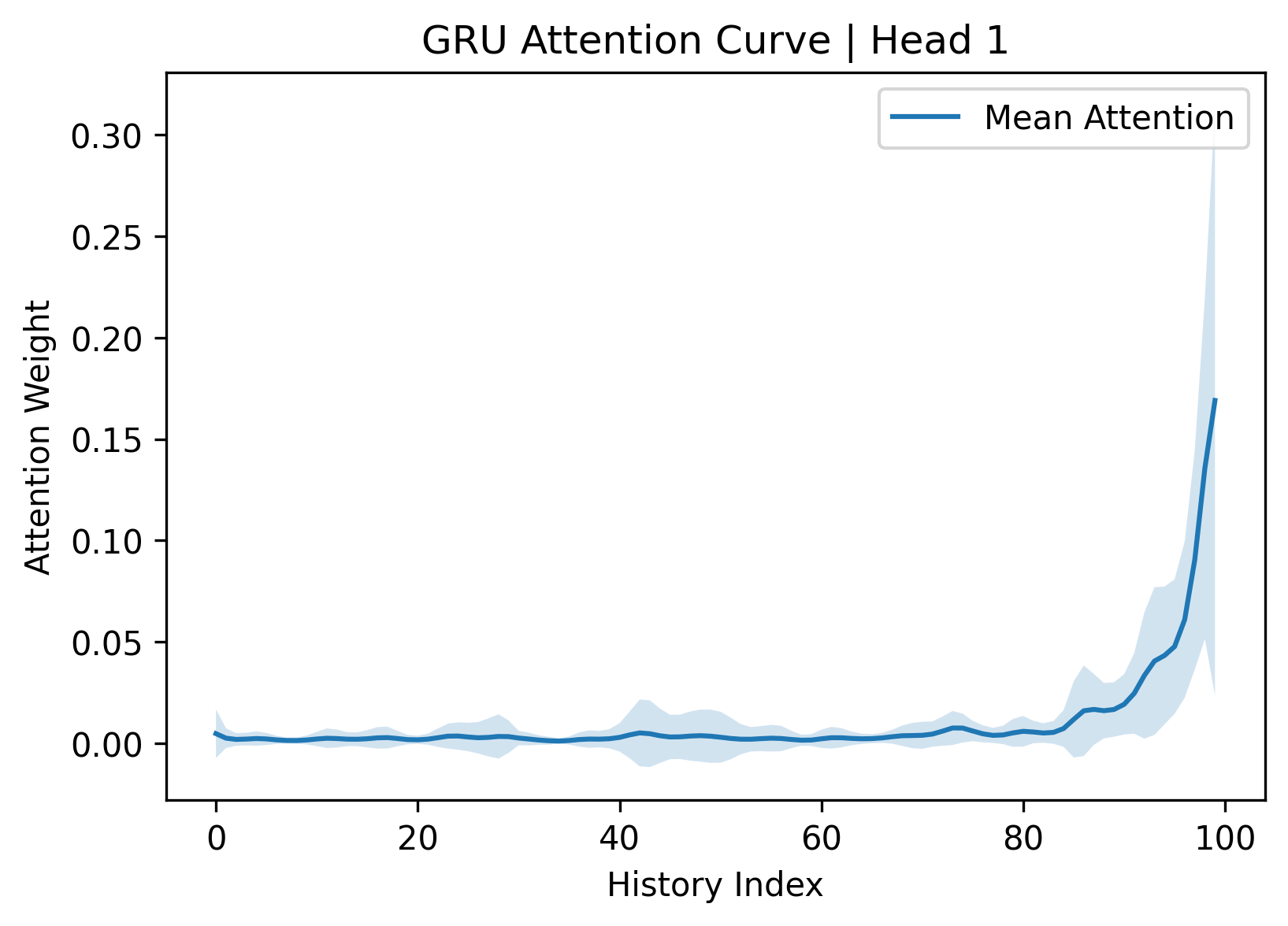} \\

\bottomrule
\end{tabular}
\vspace{3pt}
\caption{Mean GRU Attention Curves for \textbf{Stair Terrain} (17cm) Across Robots and Selected Heads}
\label{fig:stair_attention_gru}

\end{figure}
\begin{figure}[ht]
\centering
\begin{tabular}{lcc}
\toprule
\textbf{Robot $\backslash$ Head} & \textbf{Head 0} & \textbf{Head 1} \\
\midrule

\textbf{Go1 (Small)} &
\includegraphics[width=0.28\textwidth]{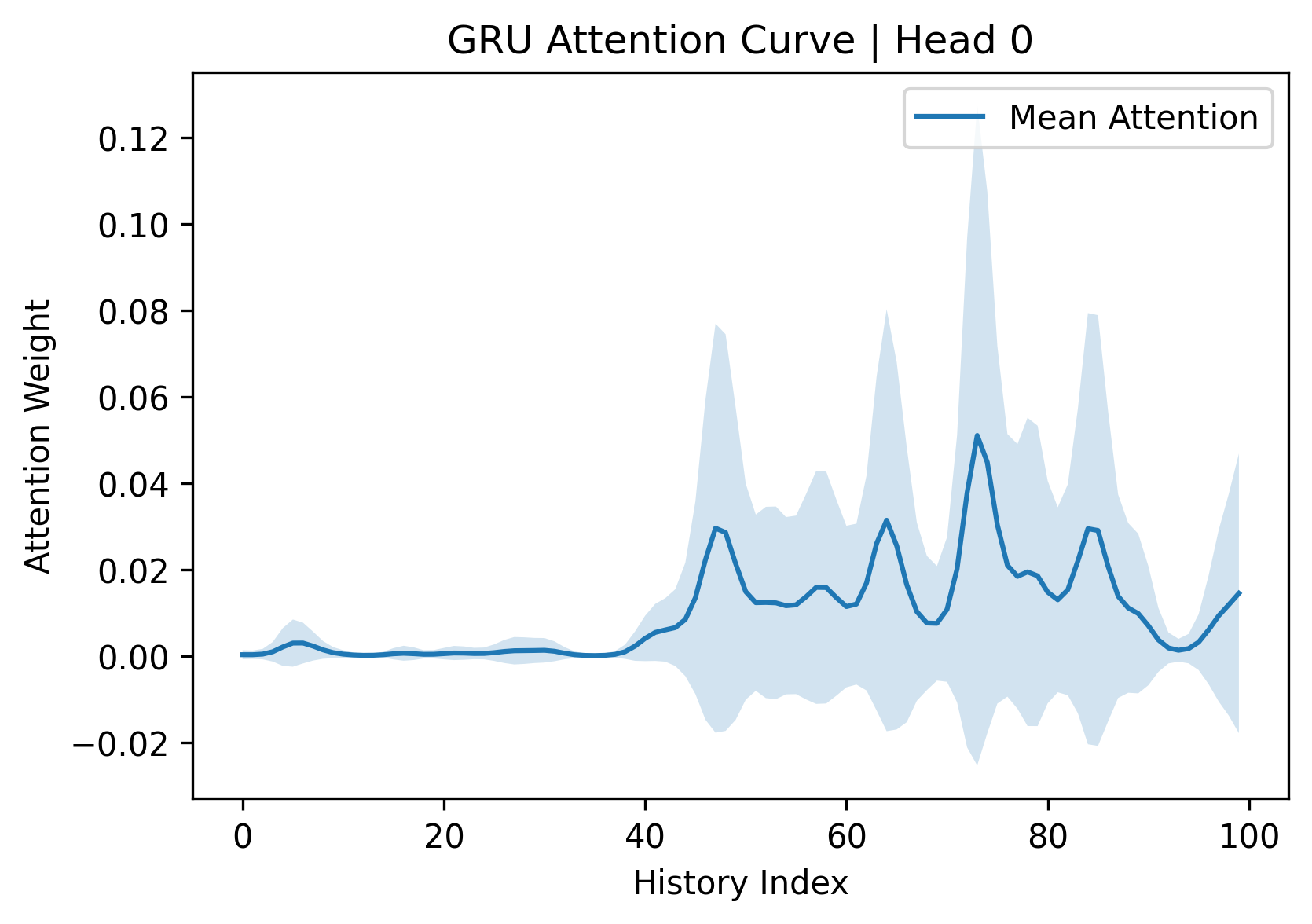} &
\includegraphics[width=0.28\textwidth]{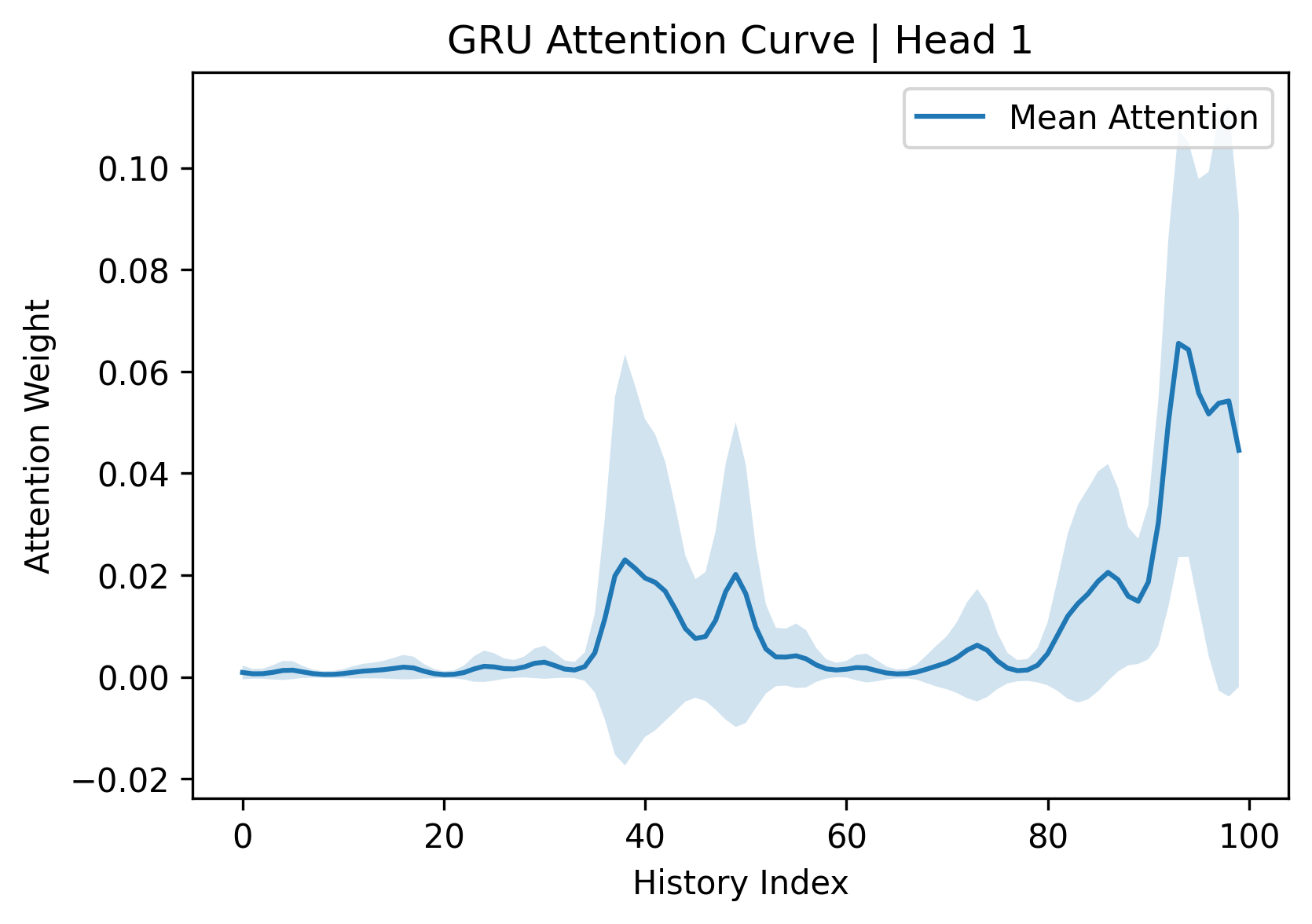} \\

\textbf{Stoch3 (Medium)} &
\includegraphics[width=0.28\textwidth]{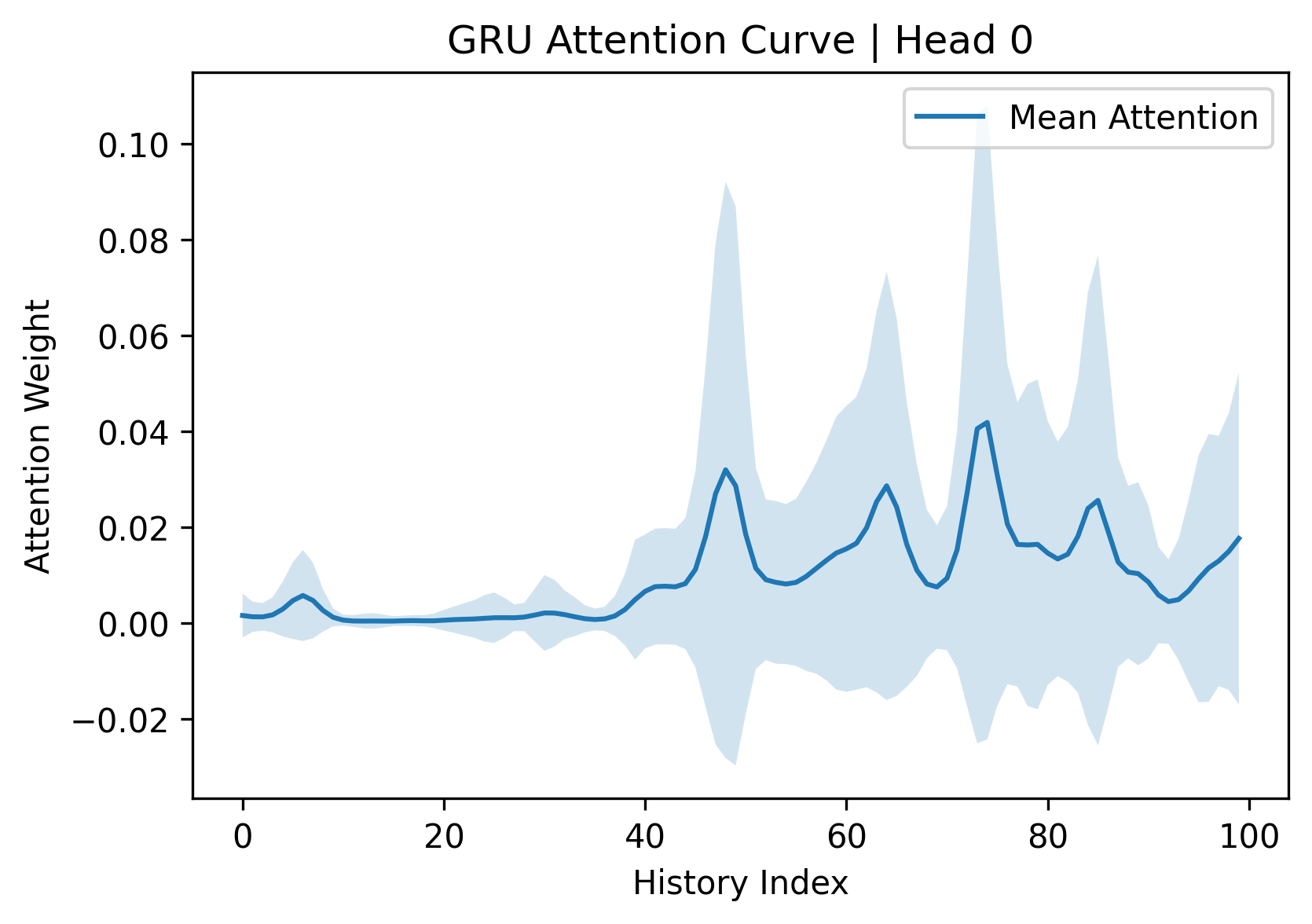} &
\includegraphics[width=0.28\textwidth]{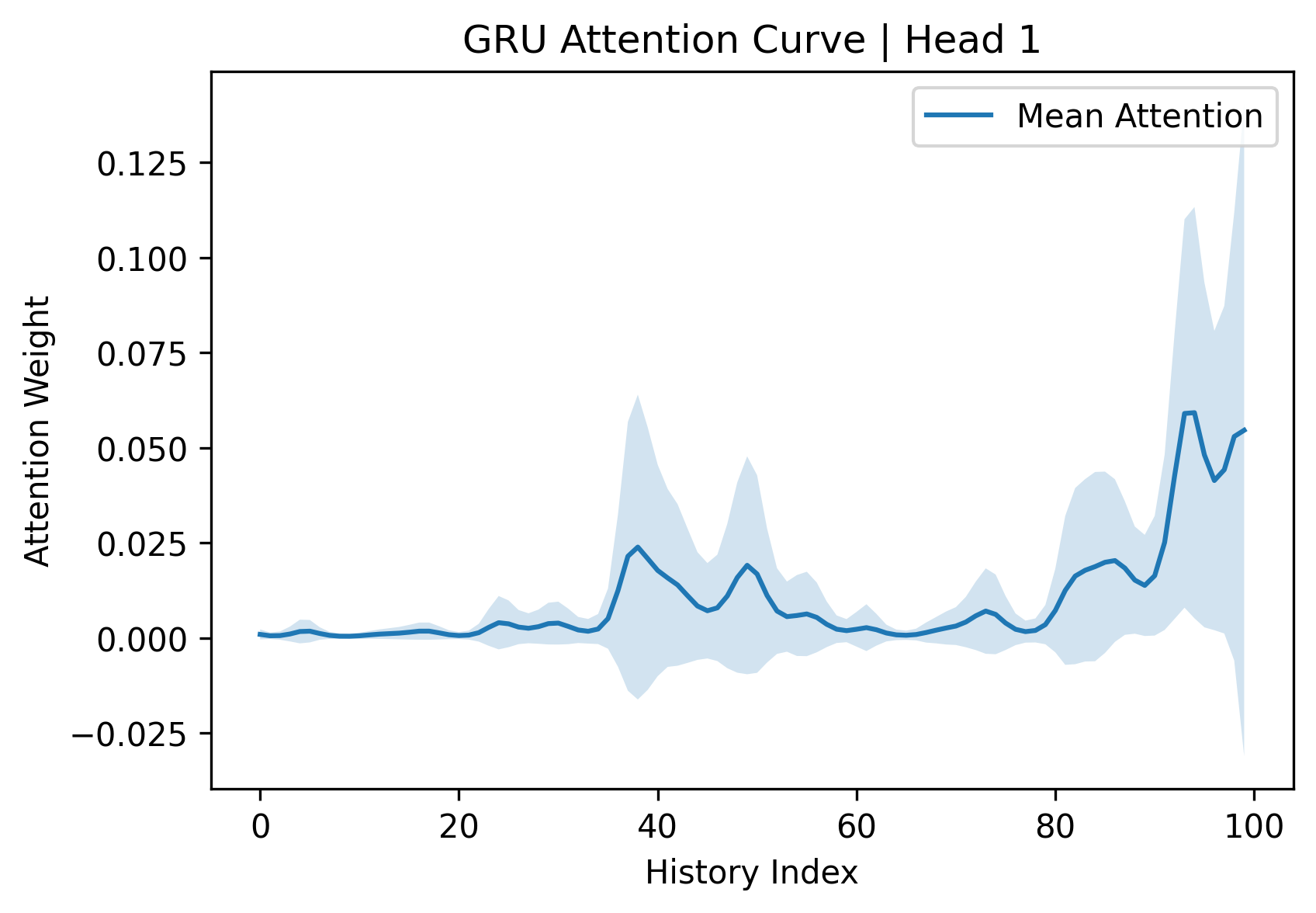} \\

\textbf{B1 (Large)} &
\includegraphics[width=0.28\textwidth]{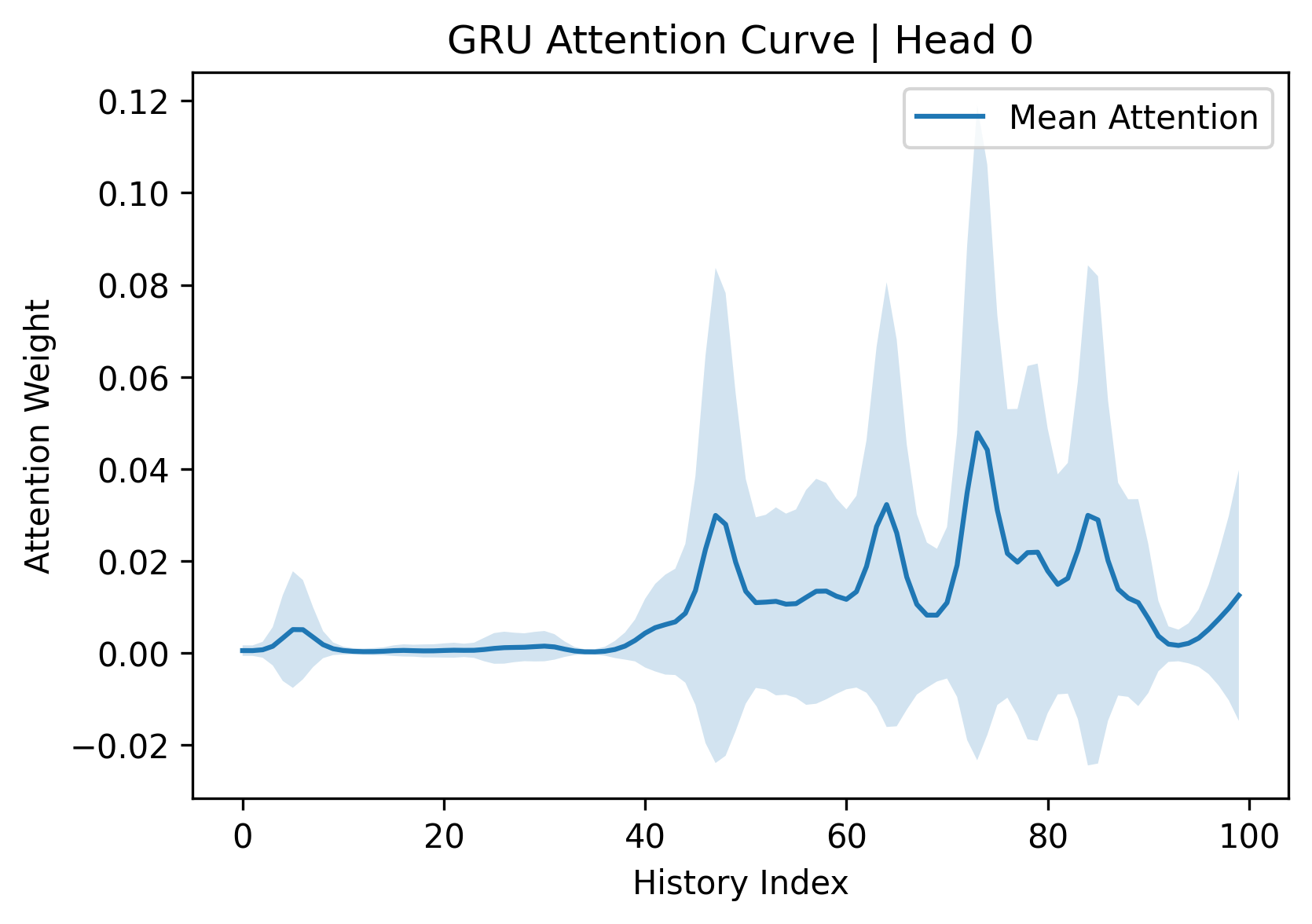} &
\includegraphics[width=0.28\textwidth]{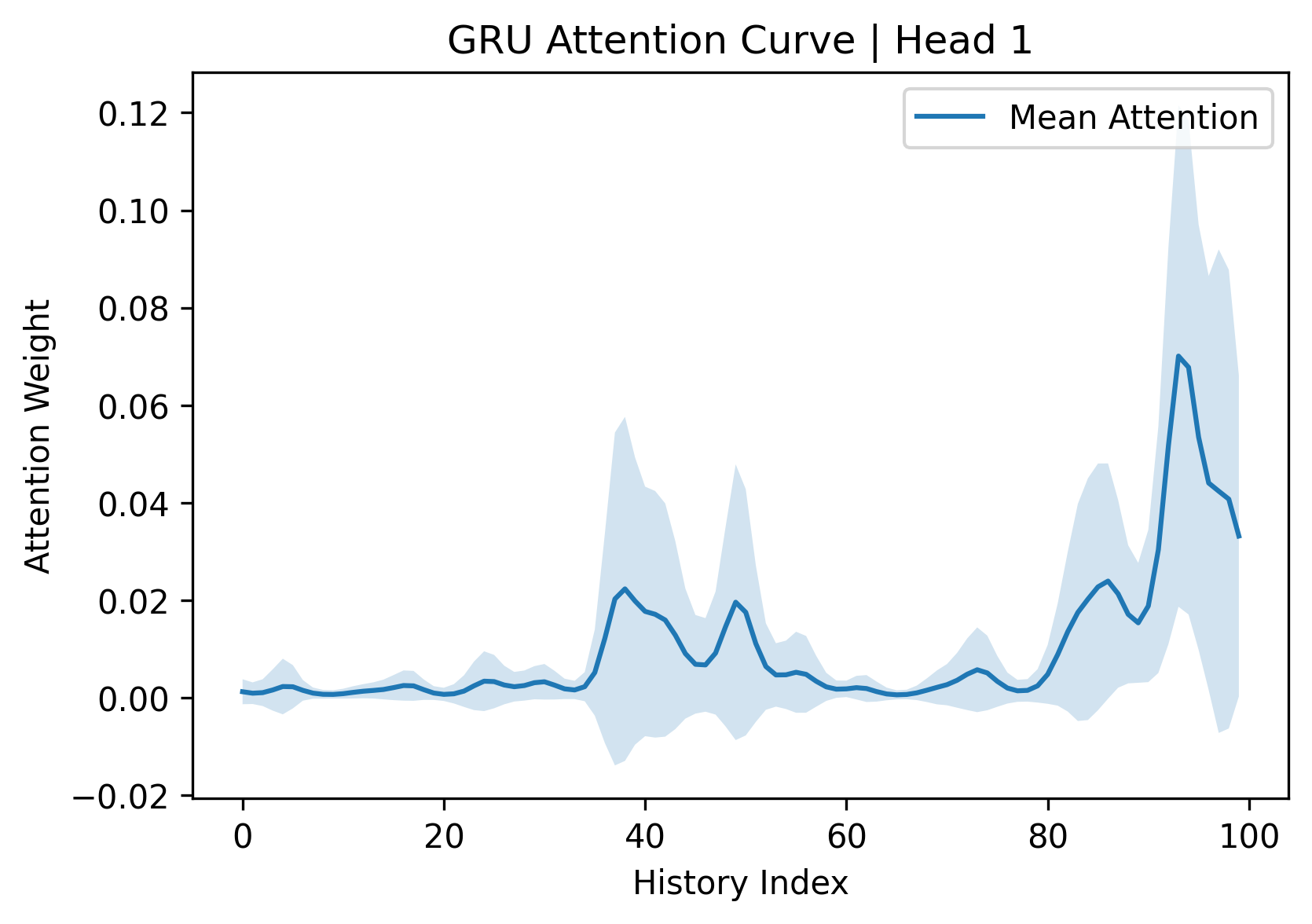} \\

\bottomrule
\end{tabular}
\vspace{3pt}
\caption{Mean GRU Attention Curves for \textbf{Slope Terrain} (Rough Slopes - 25$^\circ$) Across Robots and Selected Heads}
\label{fig:rough_slope_attention_gru}
\end{figure}
\begin{figure}[ht]
\centering
\begin{tabular}{lcc}
\toprule
\textbf{Robot $\backslash$ Head} & \textbf{Head 0} & \textbf{Head 1} \\
\midrule

\textbf{Go1 (Small)} &
\includegraphics[width=0.28\textwidth]{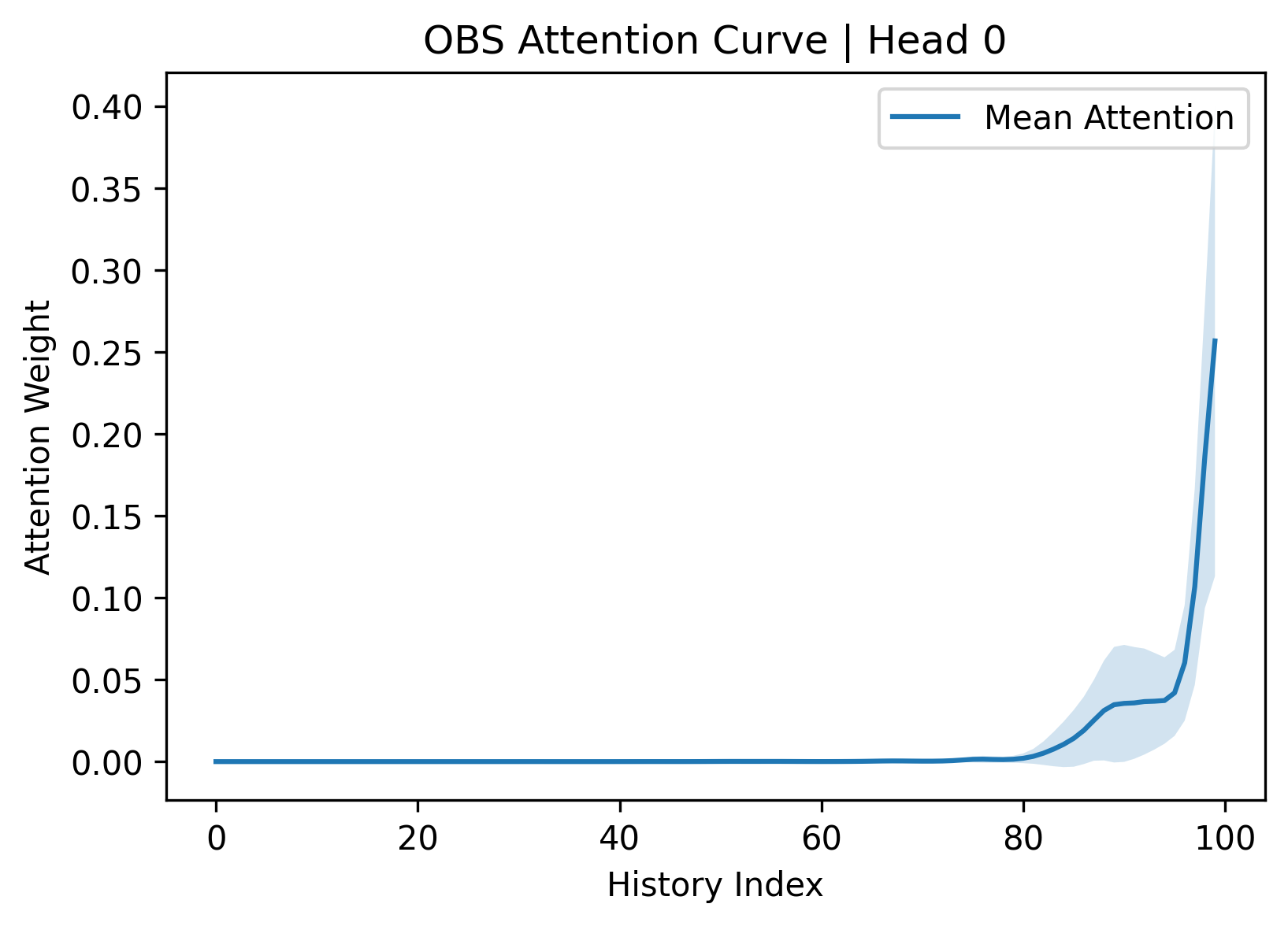} &
\includegraphics[width=0.28\textwidth]{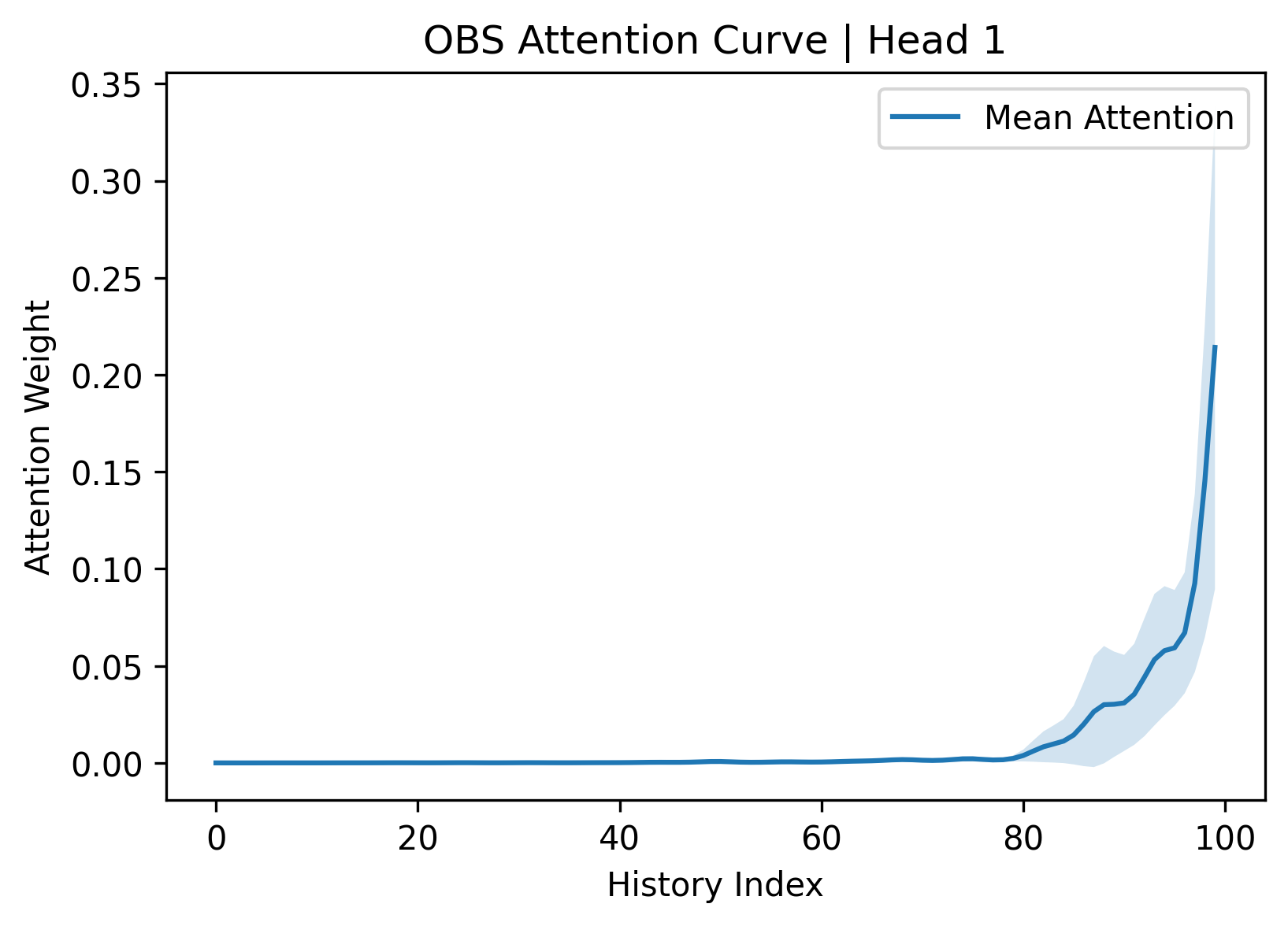} \\

\textbf{Stoch3 (Medium)} &
\includegraphics[width=0.28\textwidth]{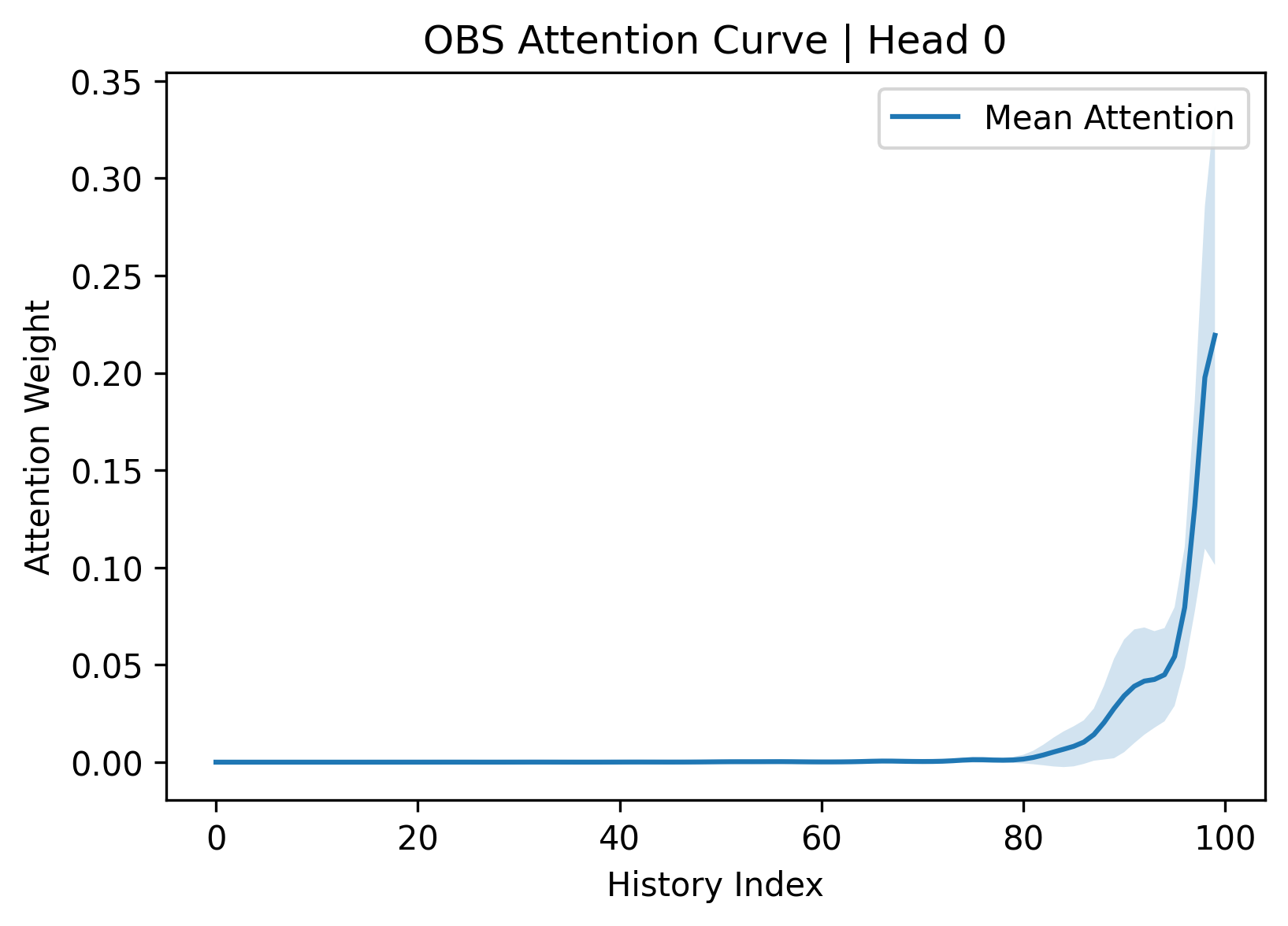} &
\includegraphics[width=0.28\textwidth]{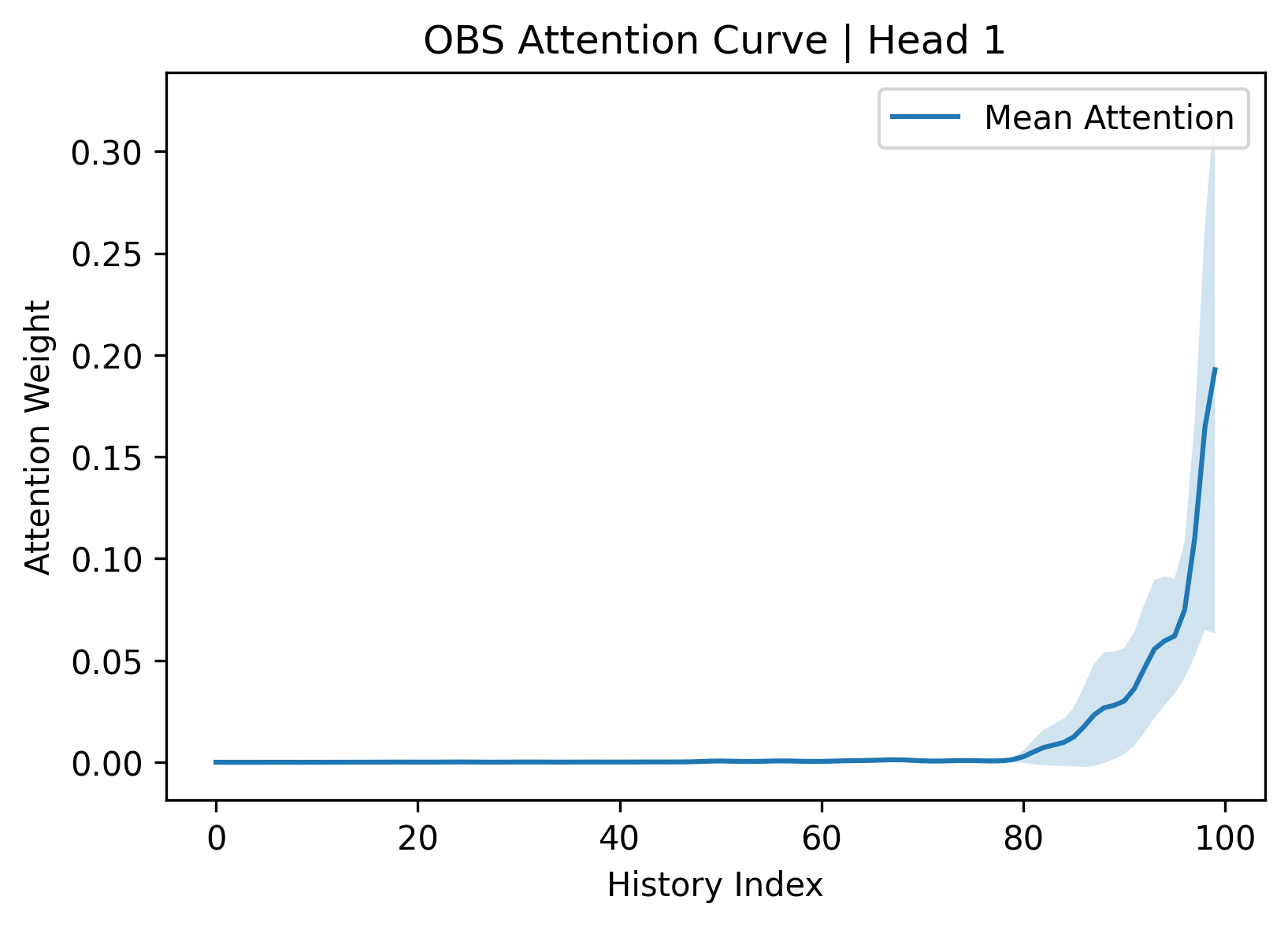} \\

\textbf{B1 (Large)} &
\includegraphics[width=0.28\textwidth]{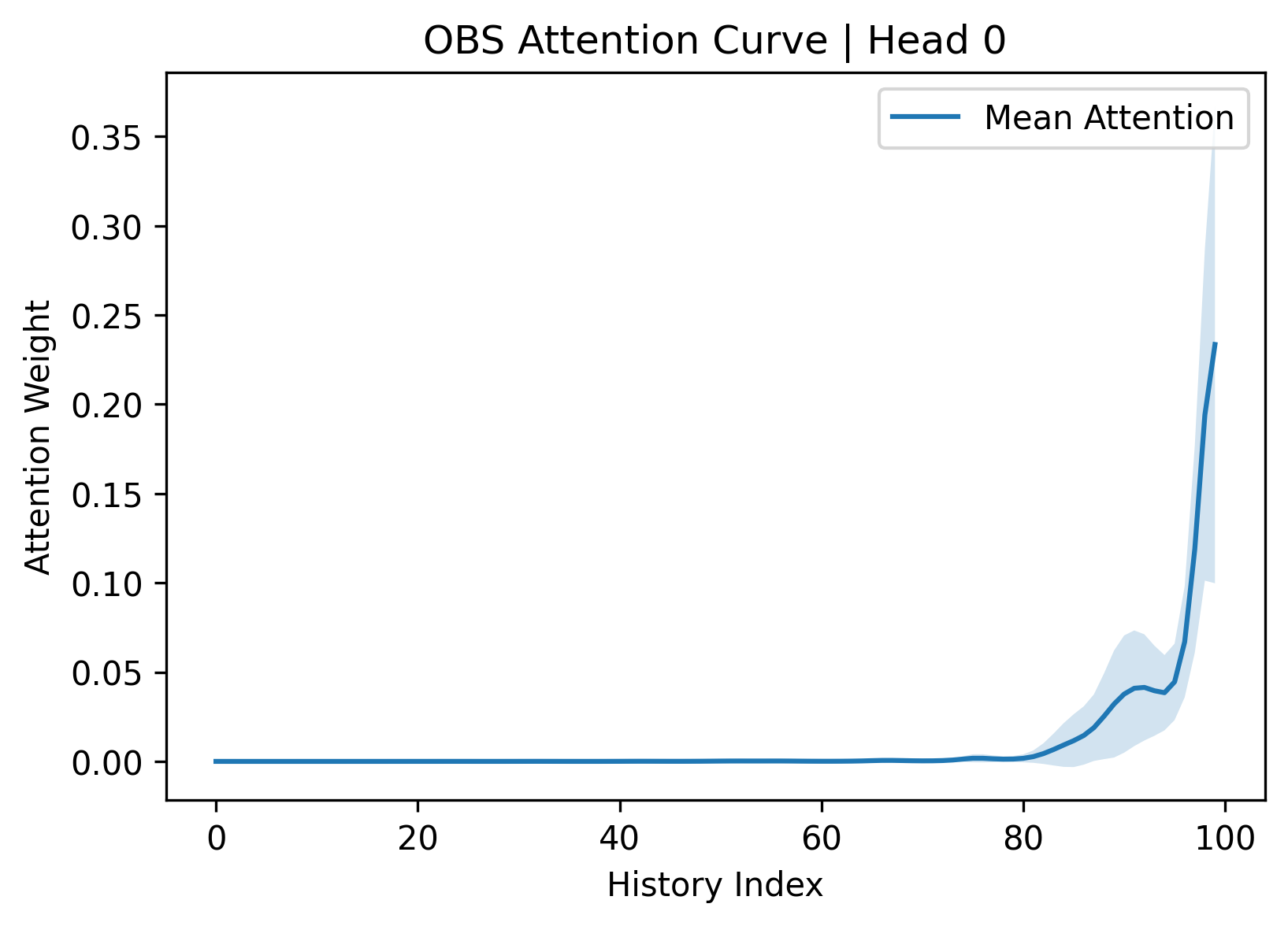} &
\includegraphics[width=0.28\textwidth]{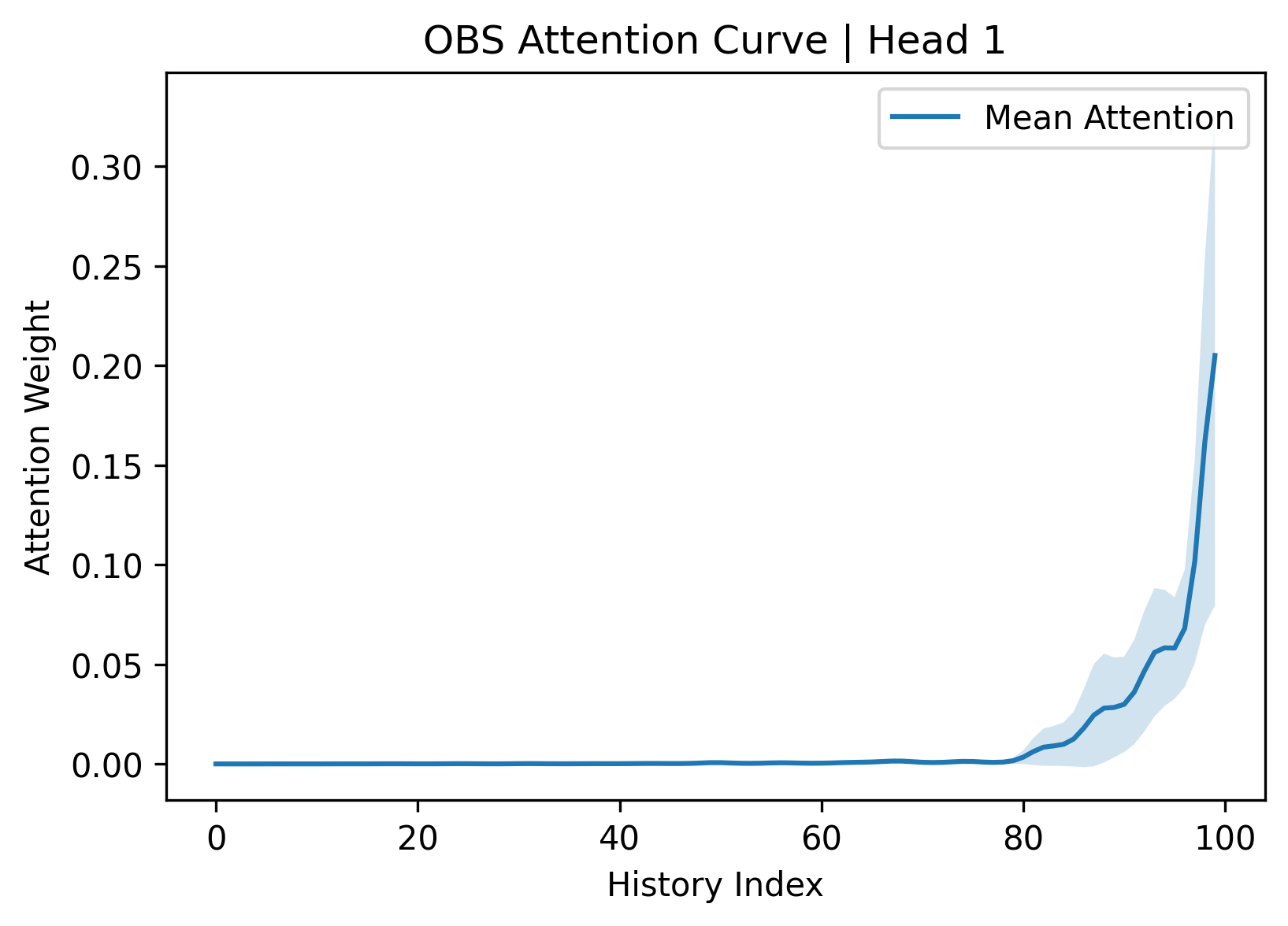} \\

\bottomrule
\end{tabular}
\vspace{3pt}
\caption{Mean OBS Attention Curves for \textbf{Flat Terrain} Across Robots and Selected Heads}
\label{fig:flat_attention_obs}
\end{figure}
\begin{figure}[ht]
\centering
\begin{tabular}{lcc}
\toprule
\textbf{Robot $\backslash$ Head} & \textbf{Head 0} & \textbf{Head 1} \\
\midrule

\textbf{Go1 (Small)} &
\includegraphics[width=0.28\textwidth]{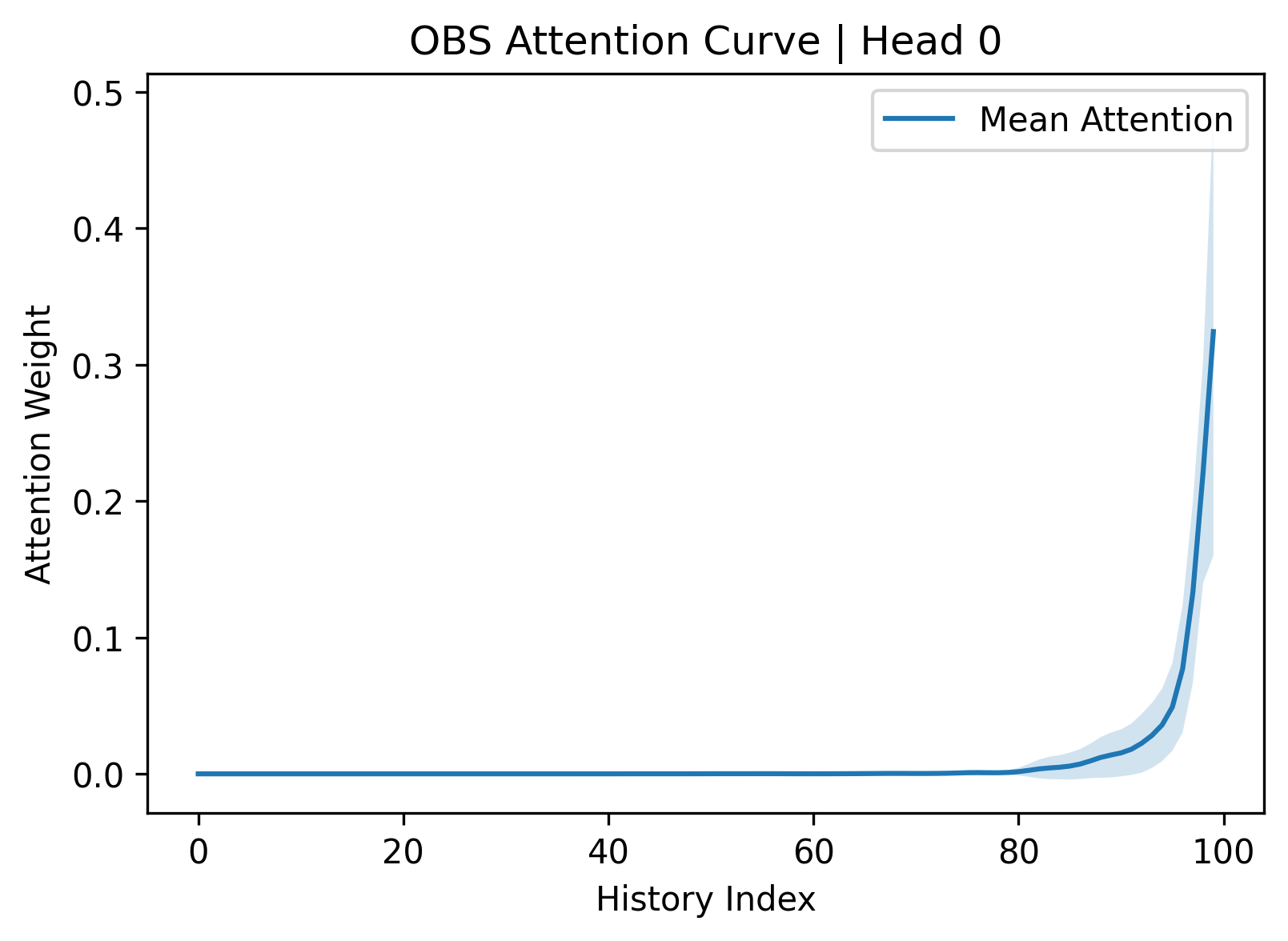} &
\includegraphics[width=0.28\textwidth]{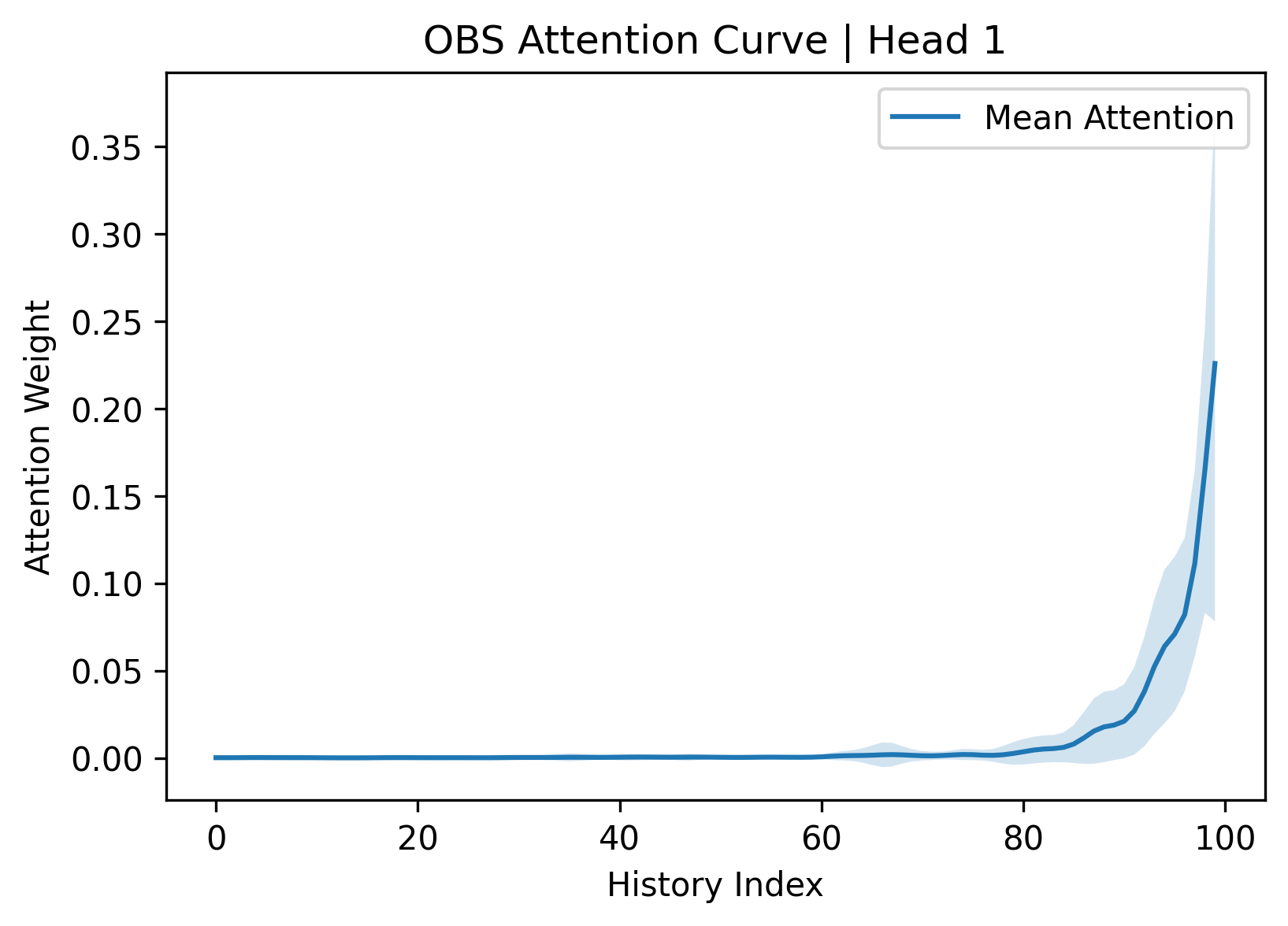} \\

\textbf{Stoch3 (Medium)} &
\includegraphics[width=0.28\textwidth]{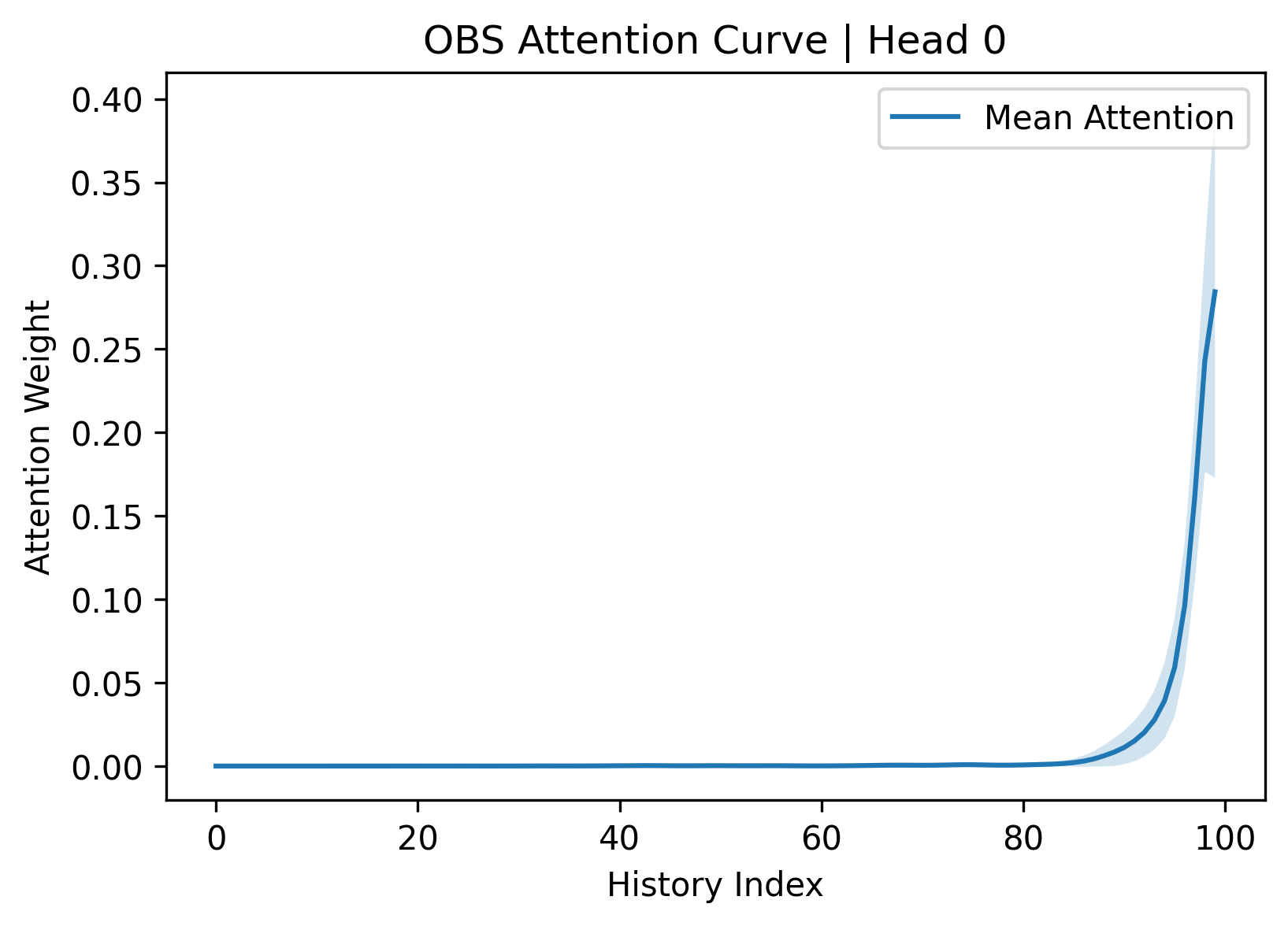} &
\includegraphics[width=0.28\textwidth]{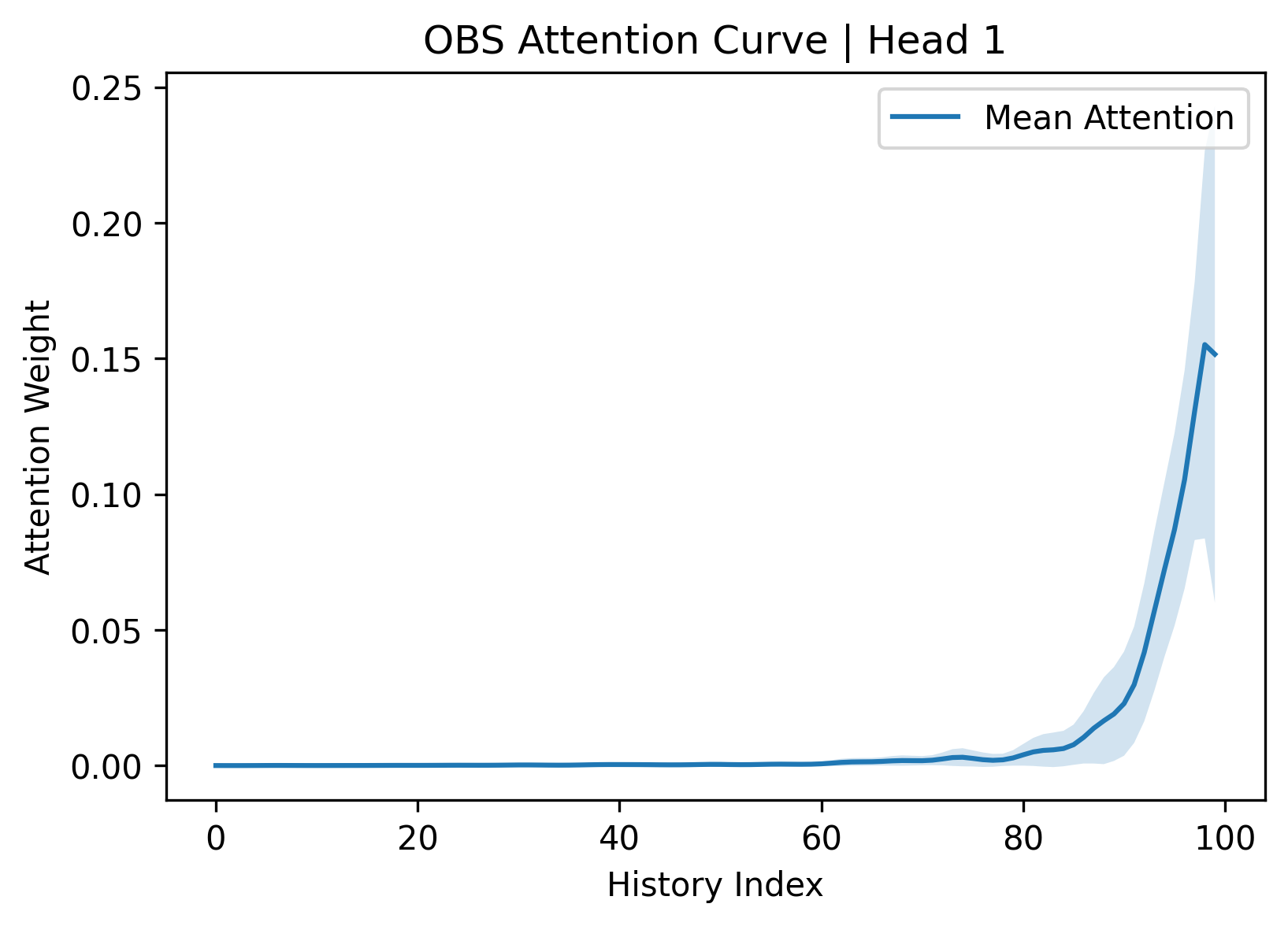} \\

\textbf{B1 (Large)} &
\includegraphics[width=0.28\textwidth]{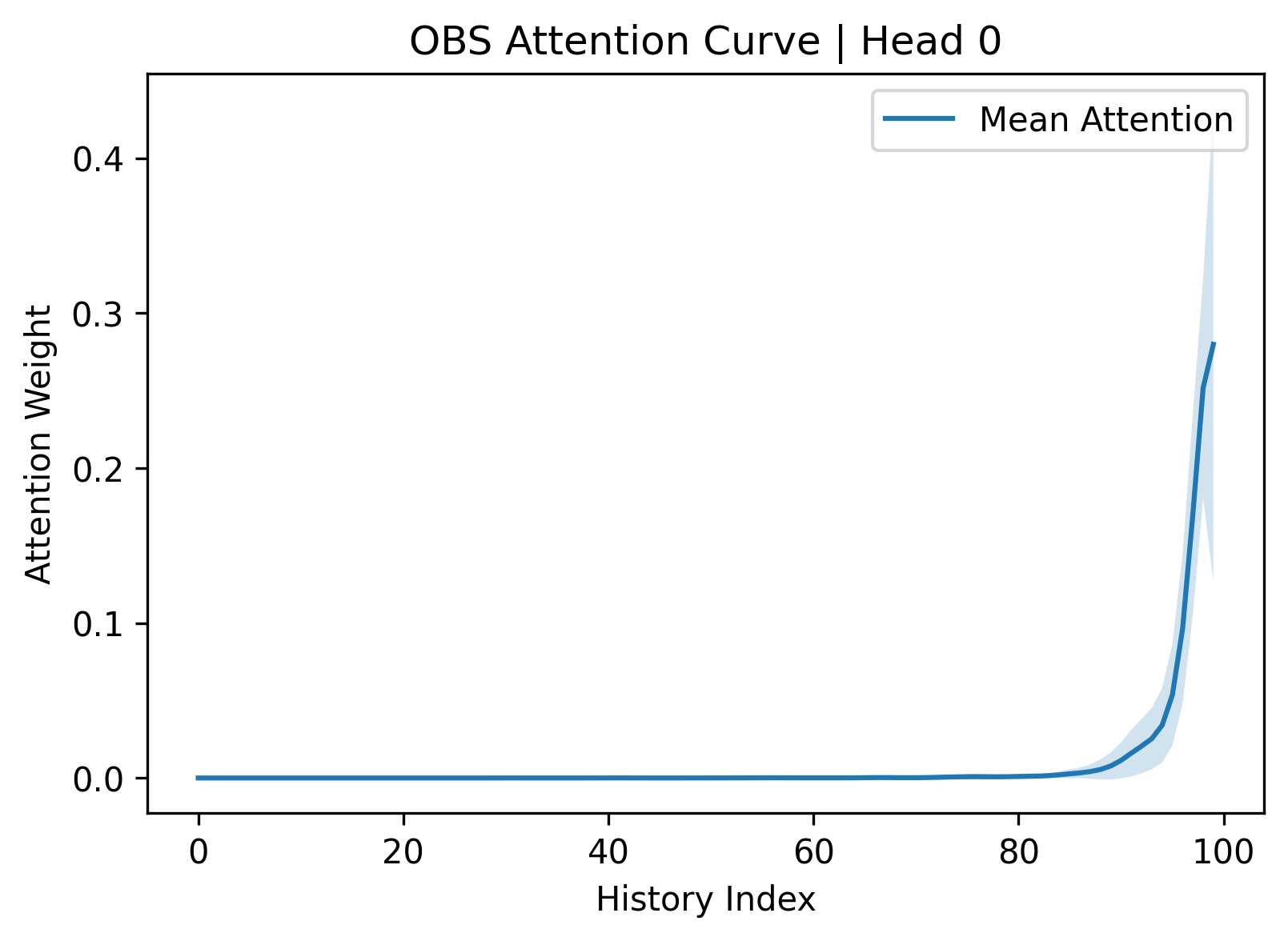} &
\includegraphics[width=0.28\textwidth]{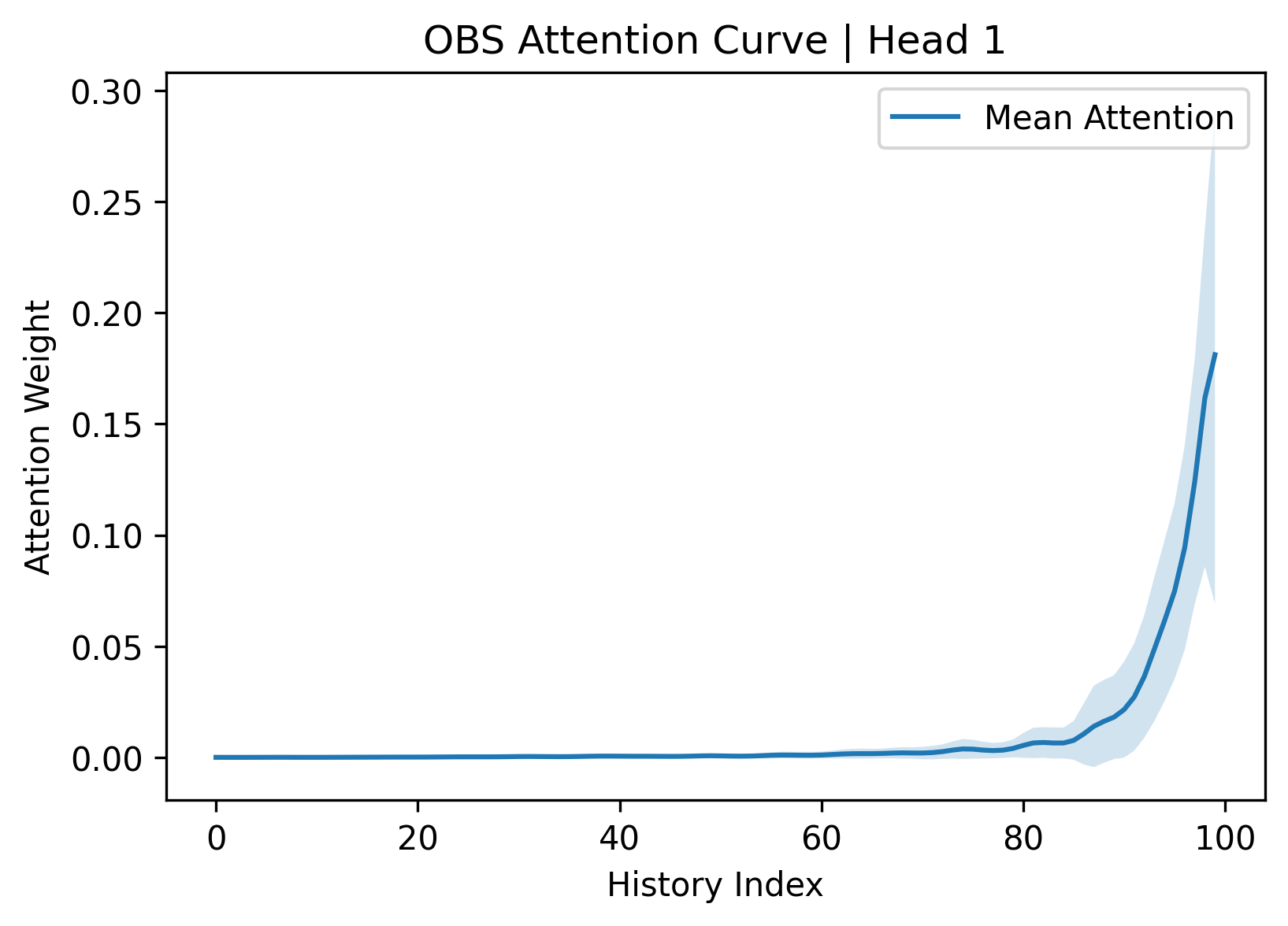} \\

\bottomrule
\end{tabular}
\vspace{3pt}
\caption{Mean OBS Attention Curves for \textbf{Stair Terrain} Across Robots and Selected Heads}
\label{fig:stair_attention_obs}
\end{figure}
\begin{figure}[ht]
\centering
\begin{tabular}{lcc}
\toprule
\textbf{Robot $\backslash$ Head} & \textbf{Head 0} & \textbf{Head 1} \\
\midrule

\textbf{Go1 (Small)} &
\includegraphics[width=0.28\textwidth]{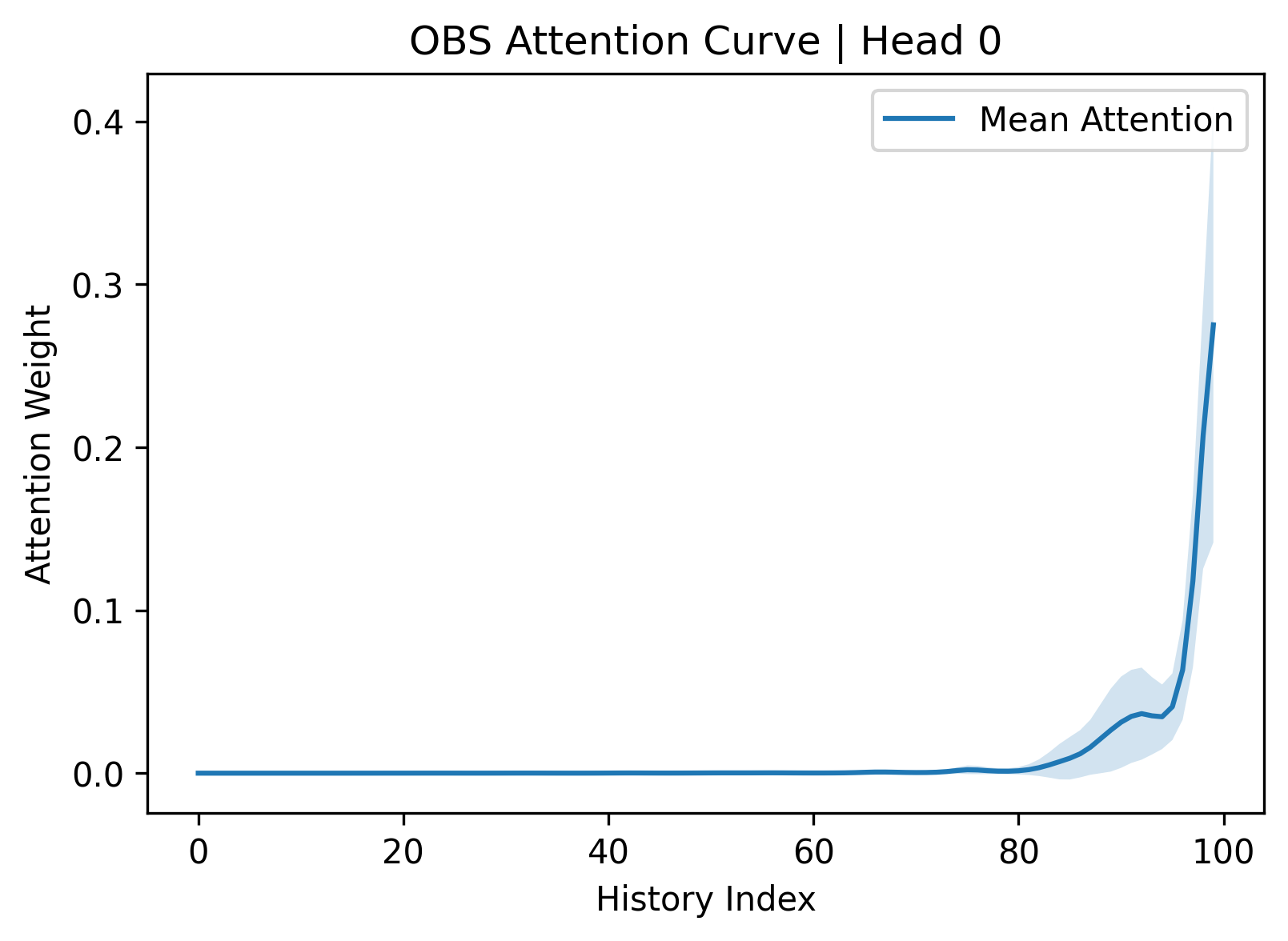} &
\includegraphics[width=0.28\textwidth]{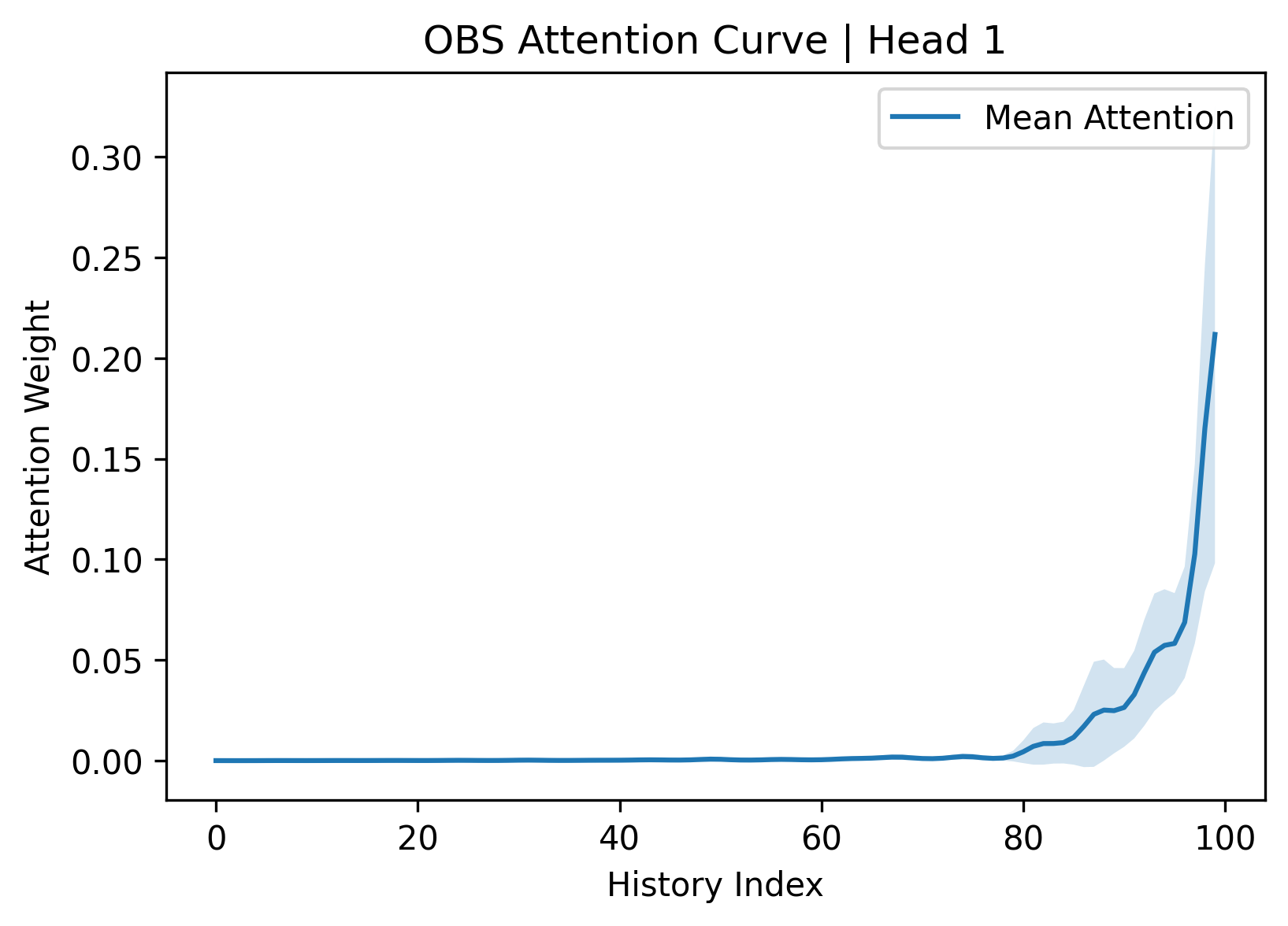} \\

\textbf{Stoch3 (Medium)} &
\includegraphics[width=0.28\textwidth]{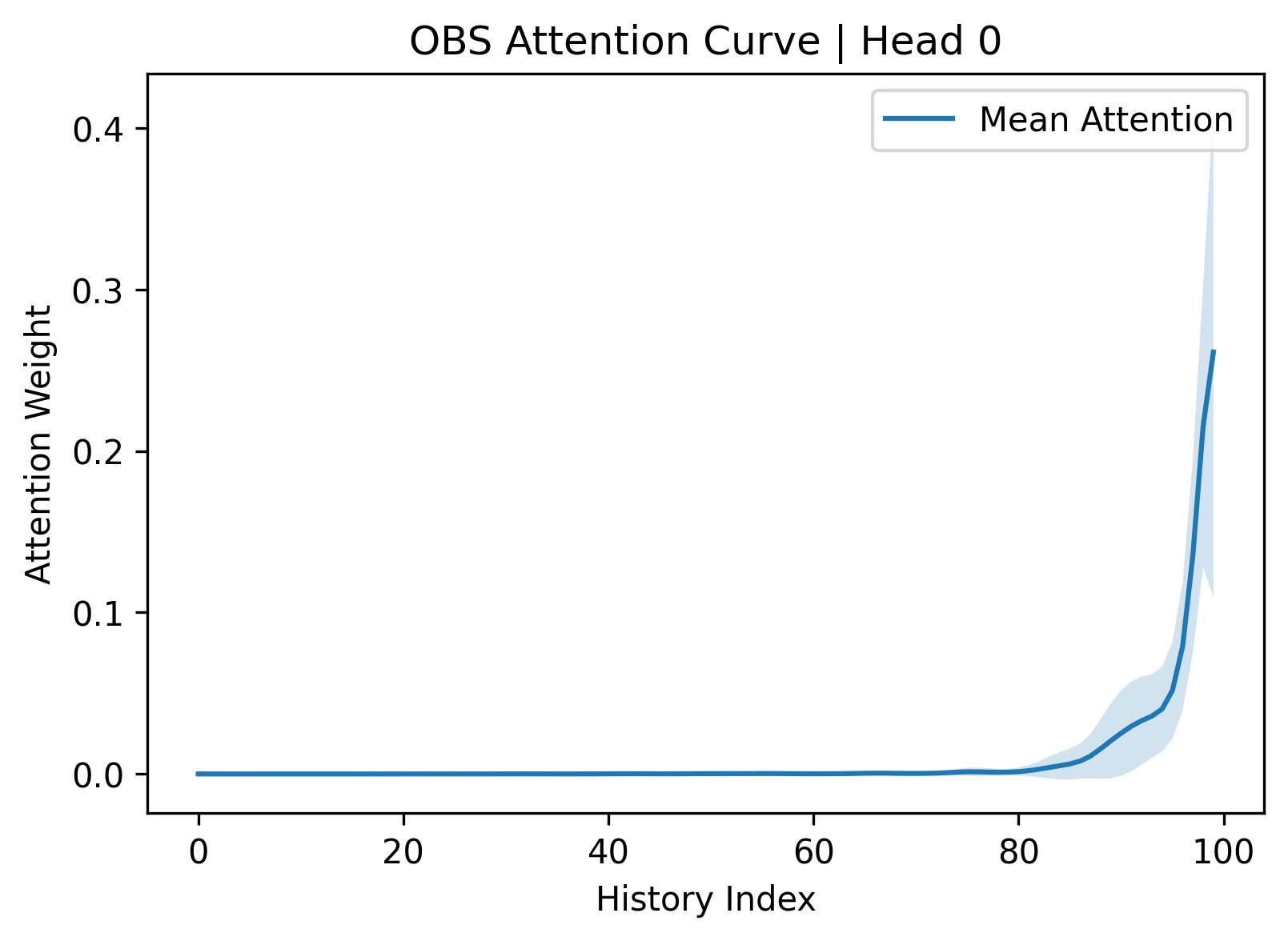} &
\includegraphics[width=0.28\textwidth]{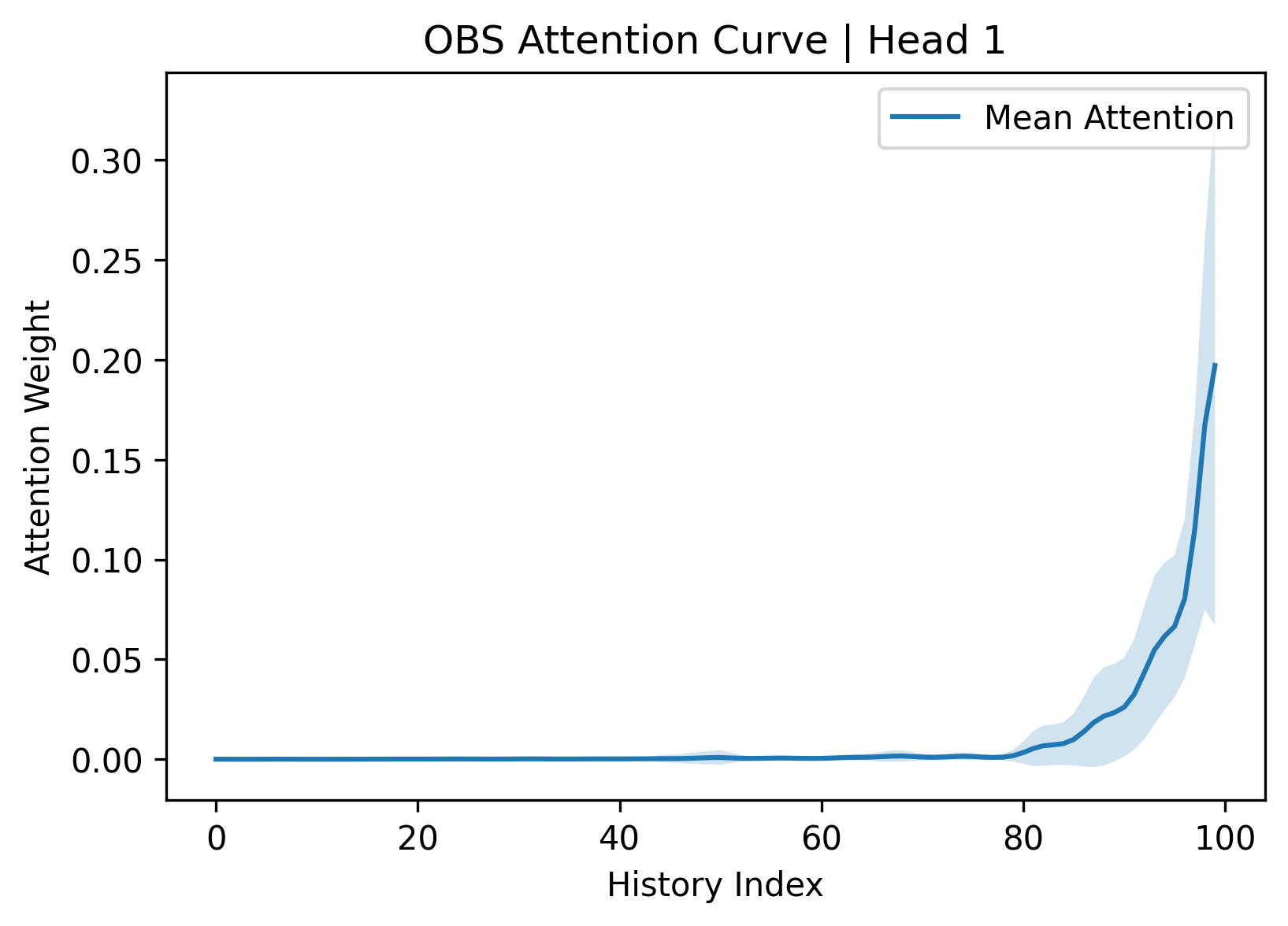} \\

\textbf{B1 (Large)} &
\includegraphics[width=0.28\textwidth]{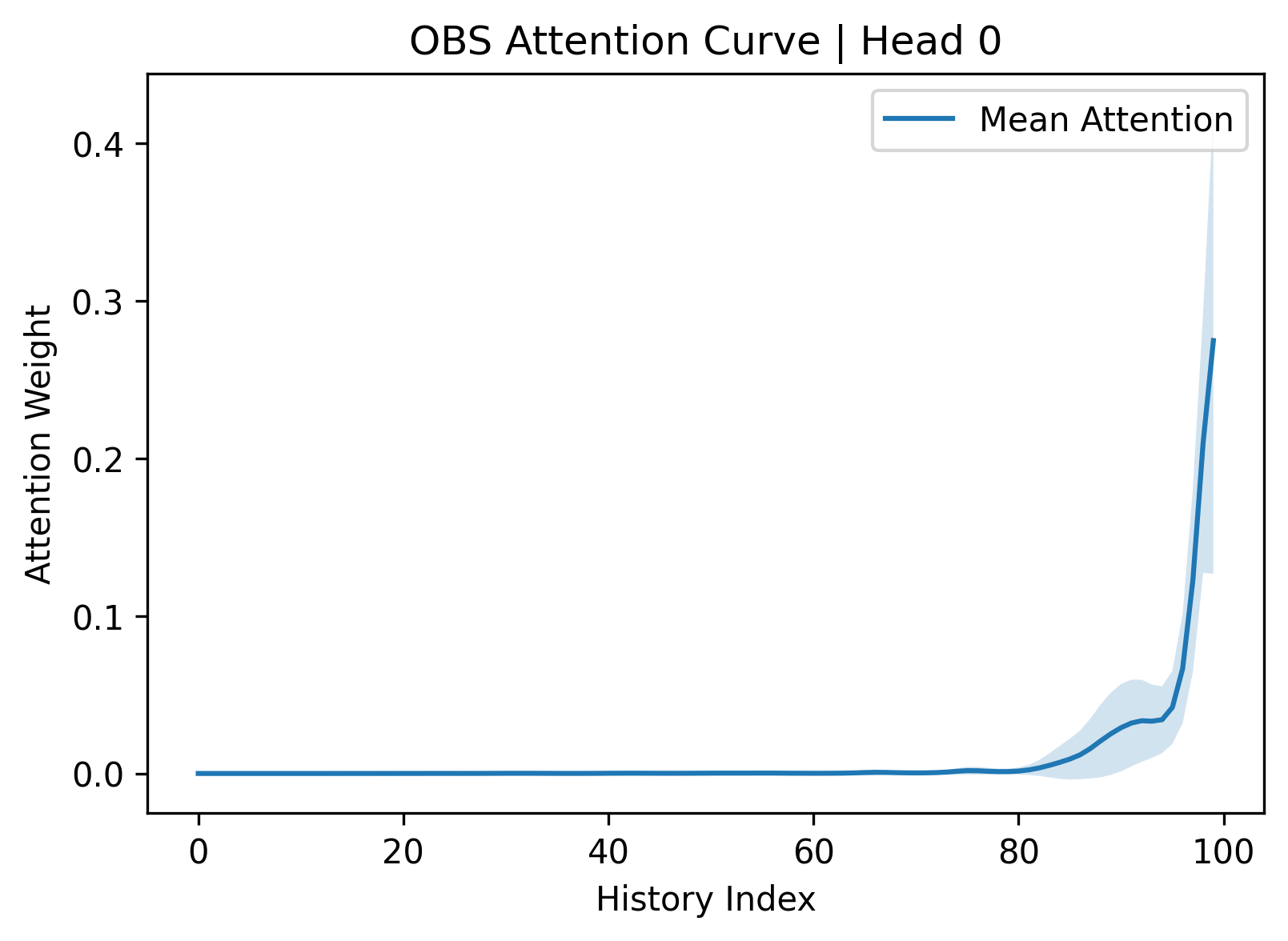} &
\includegraphics[width=0.28\textwidth]{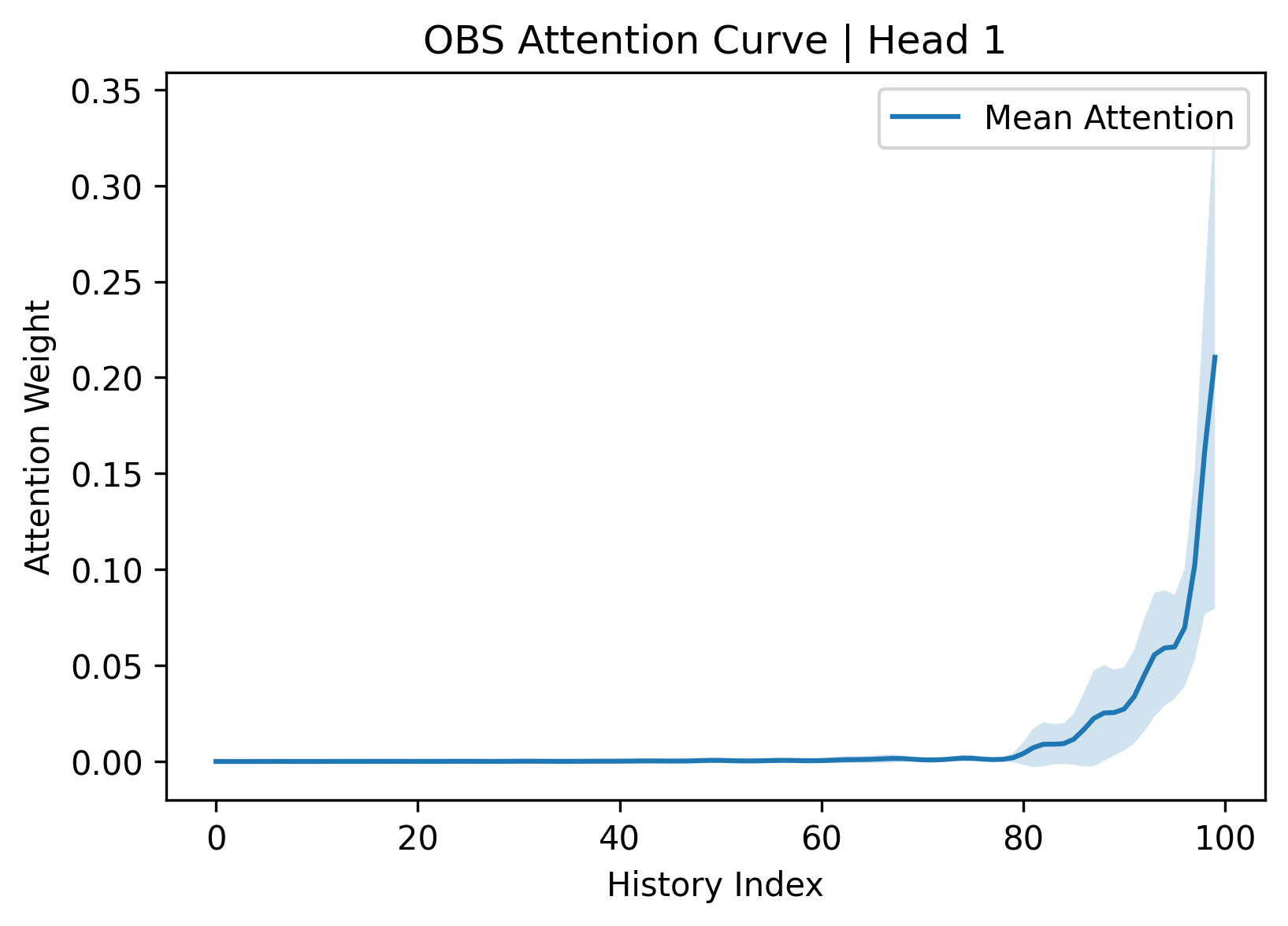} \\

\bottomrule
\end{tabular}
\vspace{3pt}
\caption{Mean OBS Attention Curves for \textbf{Slope Terrain} (Rough Slopes - 25$^\circ$) Across Robots and Selected Heads}
\label{fig:rough_slope_attention_obs}
\end{figure}

To understand how the policy attends to information during decision-making, we analyze the  mean attention scores per head of both GRU-based temporal attention and observation (OBS) attention layers. Figures \ref{fig:flat_attention_gru}, \ref{fig:stair_attention_gru}, and \ref{fig:rough_slope_attention_gru} show attention patterns for GRU embeddings across three terrains: flat, stairs, and slope. Corresponding observation attention patterns are shown in Figures \ref{fig:flat_attention_obs}, \ref{fig:stair_attention_obs}, and \ref{fig:rough_slope_attention_obs}.

\textbf{Key Insight:} Each attention head exhibits consistent patterns across different robots when conditioned on the same terrain, indicating that the policy has learned to focus on terrain-specific dynamics rather than robot-specific features.

\textbf{GRU Attention:} GRU attention heads show rhythmic and alternating patterns over time, with the attention weights oscillating between recent and earlier GRU embeddings. This reflects temporal reasoning and the use of history to maintain gait periodicity, especially in flat and slope terrains. The average attention maps confirm that previous time steps are actively attended to, suggesting the network is leveraging temporal memory to drive locomotion.

\textbf{Observation Attention:} In contrast, observation attention heads primarily focus on the most recent observation, regardless of terrain type, which could indicate that immediate sensory feedback is critical for terrain-reactive behavior.

\textbf{Robots Used for Evaluation:}
\begin{itemize}
    \item \textbf{Go1 (small)} — Agile and lightweight
    \item \textbf{Stoch3 (medium)} — Mid-weight, versatile platform
    \item \textbf{B1 (large)} — Heavy-duty robot for large terrain disturbances
\end{itemize}

Overall, the shared attention behavior across robot morphologies reinforces the terrain-conditioned generalization capability of the policy, validating the design's robustness and scalability.

\end{document}